\documentclass[twocolumn]{article}
\usepackage[english]{babel}
\usepackage[nolist]{acronym}
\usepackage{subcaption}
\usepackage[T1]{fontenc}
\usepackage[english]{babel}
\usepackage{tabularx}
\usepackage{multirow}
\usepackage{booktabs}   
\usepackage{amssymb} 
\usepackage[a4paper, total={6in, 8in}, margin=0.680in]{geometry}
\usepackage{float}
\providecommand{\BIBforeignlanguage}[2]{#2}
\usepackage{amsmath}
\usepackage{graphicx}
\usepackage{wrapfig}
\usepackage[colorlinks=true, allcolors=blue]{hyperref}

\begin{document}
\twocolumn[{
    \begin{center}
        {\LARGE\bfseries Energy Overhead in Sensor Nodes: Embedded Machine Learning and Data Offloading via Low Power Wide Area Networks\par}
        \vspace{1em}
        {\large Benjamin Karic, Nina Herrmann, Jan Stenkamp, Paula Scharf, Fabian Gieseke and Angela Schwering\par}
    \end{center}

    \small 
    \begin{minipage}[t]{0.525\textwidth}
    \noindent\textbf{Abstract}---
    Image- and audio-based remote IoT applications offer a substantial opportunity to address ecological changes. However, to be effective, they must operate for long periods on limited power while also providing data via LPWAN. Previous studies have investigated the energy efficiency of isolated components such as radio transmission or data processing methods within an IoT system. However, a holistic evaluation is required to determine the main drivers of energy consumption in sensor nodes. To address this, we systematically analyze energy consumption across the entire sensor-node pipeline, from data acquisition to optional on-device CNN classification and wireless network transmission, using an ESP32-S3 microcontroller. We compare on-device CNN classification followed by result transmission with transmitting raw sensor data across three environmental monitoring scenarios.    
    \end{minipage}%
    \hfill
    \begin{minipage}[t]{0.435\textwidth}
    \vspace{0pt} 
    \centering
    \includegraphics[width=\linewidth]{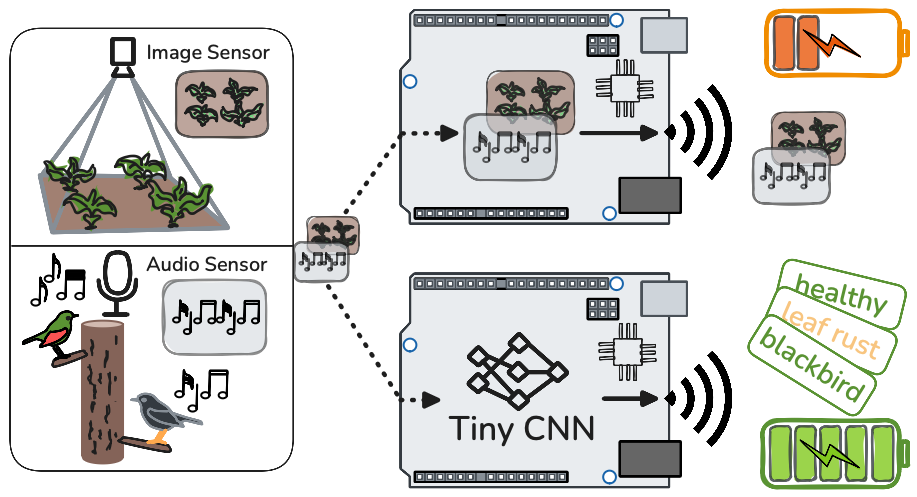}
    \label{fig:visual-abstract}
    \end{minipage}
    Thereby, we vary developer-controllable variables, notably the model architecture and the LPWAN technologies (LTE-M, NB-IoT, LoRaWAN). Our results indicate that on-device inference can reduce sensor node energy consumption by 1.7x to 6.9x compared to raw data transmission, depending on transmission frequency, with an acceptable loss in accuracy.
    Further, our results narrow down the energy tradeoff to variables IoT developers can control. Therefore, our findings provide empirical guidance for designing energy-efficient, long-range sensing systems and demonstrate that TinyML enables sustainable, autonomous deployments. 
    \textbf{Note:} This work has been submitted to the IEEE for possible publication. Copyright may be transferred without notice, after which this version may no longer be accessible.
    \par
    \vspace{2em}
}]

\begin{acronym}[ECU]
\acro{iot}[IoT]{Internet of Things}
\acro{ai}[AI]{artificial intelligence}
\acro{cnn}[CNN]{convolutional neural network}
\acrodefplural{cnn}[CNNs]{convolutional neural networks}
\acro{EmML}[EmbeddedML]{embedded machine learning}
\acro{tinyml}[TinyML]{tiny machine learning}
\acro{ml}[ML]{Machine Learning}
\acro{mcu}[MCU]{microcontroller unit}
\acro{squeezenet}[SqueezeNet]{SqueezeNet}
\acro{mcunet}[MCUNet]{MCUNet}
\acro{mobilenet}[MobileNetV2]{MobileNetV2}
\acro{ptq}[PTQ]{post-training quantization}
\acro{qat}[QAT]{quantization-aware training}
\acro{cub}[CUB]{Caltech-UCSD Birds-200-2011}
\acro{plantvillage}[PlantVillage]{PlantVillage}
\acro{yellowhammer}[Yellowhammer]{Yellowhammer bird vocalizations}
\acro{eDRX}[eDRX]{Extended Discontinuous Reception}
\acro{PSM}[PSM]{Power Saving Mode}
\acro{mfcc}[MFCC]{mel-frequency cepstral coefficients}
\acro{SoC}[SoC]{System on a Chip}
\acro{tcresnet}[TCResNet8]{TCResNet8}
\acro{lorawan}[LoRaWAN]{Long Range Wide Area Network}
\acro{nbiot}[NB-IoT]{Narrowband-IoT}
\acro{ltem}[LTE-M]{Long-Term Evolution for Machines}
\acro{lpwan}[LPWAN]{low-power wide-area network}
\acro{sf}[SF]{spreading factor}
\acro{rssi}[RSSI]{received signal strength indication}
\acro{cinr}[CINR]{carrier-to-interference plus noise ratio}
\acro{rsrq}[RSRQ]{Reference Signal Received Quality}
\acro{rsrp}[RSRP]{Reference Signal Received Power}
\acro{ble}[BLE]{Bluetooth Low Energy}
\end{acronym}

\section{Introduction}

Many applications now use sensors to retrieve large amounts of data, amplifying the need for efficient sensor algorithm co-design. With such systems, however, resource use must be efficient, as energy consumption represents a key parameter that can limit or extend a system's lifetime; this is important for devices like wearables, hearing aids, and drones. 
Yet, energy efficiency is even more important for sensor systems in remote locations with limited power supply, as replacement and maintenance are costly, sometimes even decisive for system deployment, such as weather stations, acoustic recorders, and profiling floats.
These applications are typically deployed on sensor nodes, composed of one or multiple \acp{mcu}
along with sensors, communication modules, and a limited power supply. Sometimes these systems are supported with energy-harvesting techniques, but sufficient energy generation can rarely be guaranteed; these systems must save energy wherever possible.

\begin{figure*}[tb]
    \centering
    \begin{minipage}[b]{0.49\linewidth}
        \includegraphics[width=1\textwidth]{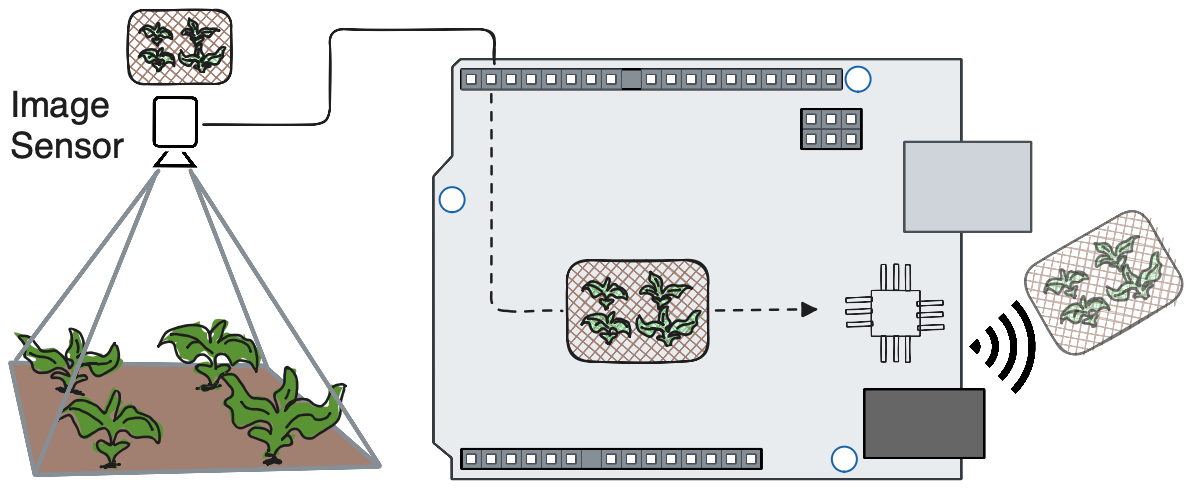}
        \subcaption{Monitoring crop health of cultivated plants
        }
        \label{fig:exampleiota}
    \end{minipage}
    \hfill
    \begin{minipage}[b]{0.49\linewidth}
        \includegraphics[width=1\textwidth]{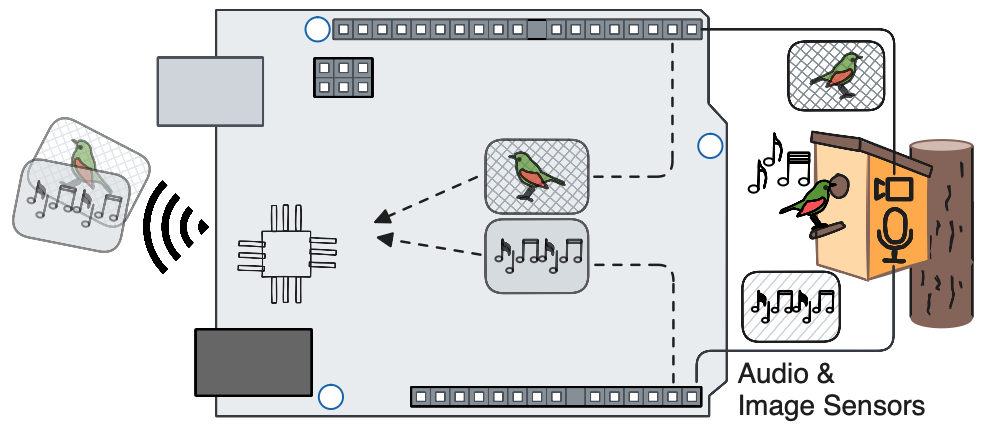}
        \subcaption{Identifying/counting birds for biodiversity monitoring}
        \label{fig:exampleiotb}
    \end{minipage}
    \caption{Examples of remote environmental monitoring applications using image and audio data and \ac{iot} concepts. Both applications usually operate in locations without a static power supply and bad connectivity.}
    \label{fig:exampleiot}
\end{figure*}
Two examples of sensor system applications for remote sensing that require energy efficiency are shown in Figure~\ref{fig:exampleiot}:
(a) observing crop health \cite{pavel2019iot} and (b) classifying (bird) species for biodiversity monitoring using cameras \cite{niers_birdf_2022} or passive acoustic monitoring \cite{shonfield2017autonomous}.
For these applications, the most energy-intensive steps involve transferring data to a server or another device
~\cite{krizanovic_advanced_2023, callebaut2018long}. Significant energy is used to initialize the transfer, and energy requirements increase with message size
~\cite{mekki2019comparative, Sinha_LPWANBIOT}. 
Thus, energy requirements can be massive for applications with high-resolution data, such as images, acoustic signals, or ECG waveforms. In such scenarios, developers can save energy and costs by sending only a reduced subset of the collected data, such as by applying compression or filter methods to reduce the data size~\cite{oehmcke2022input,stojkoska2017data,pikatkowski2023study,adaptivefilters,dynamicfilter}.
These methods often perform well on specific data patterns, such as highly repetitive sequences or temporal correlation. Still, they generalize poorly, i.e., they achieve low compression rates or introduce information loss elsewhere. 

Meanwhile, \ac{ai} methods for data analysis have been increasingly adopted in the aforementioned sensor-based monitoring applications~\cite{tzounis_internet_2017, grgic_internet_2020, rizvi_revolutionizing_2024, shoaib_advanced_2023,niers_birdf_2022}.
In the conventional approach, data are collected on the sensor system and transmitted to another, usually more powerful, device for further processing. The large amounts of data to be transferred render \acp{lpwan} with low bandwidth and duty-cycle limitations unusable and lead to high energy consumption.
With advances in AI, the rapidly growing field of \ac{EmML} now supports the creation of \textit{tiny} models that \ac{iot} developers can deploy on devices with very limited computational resources. Common model compression techniques, such as pruning and quantization, can be used to considerably reduce the memory footprint while strongly limiting the loss in prediction accuracy. 
This enables \ac{ai} methods that are usually known for their large memory footprint, such as \acp{cnn}, to be employed on resource-constrained devices. Various studies on {\ac{EmML} and \acp{cnn} have investigated trade-offs between model accuracy and size, as well as ways to reduce computational complexity to improve energy efficiency~\cite{rachmanto_characterizing_2024, muhoza_power_2023}. 

Despite this, developing a functional, energy-efficient application and selecting appropriate hardware remain challenging. While \ac{EmML} research often relies on FLOPs as a proxy for efficiency, the actual energy consumption is highly device-dependent. 
Moreover, the choice of network protocols is closely tied to the volume and structure of the data produced by the sensor or any subsequent on-device processing technique. This creates a strong interdependency between the processing unit and the radio, which cannot be understood by analyzing either component in isolation.
Therefore, further studies are required to examine individual implementation options and to characterize the energy consumption of a complete realization across its sub-tasks. In this work, we address this by systematically profiling the energy costs of data acquisition, local inference, and transmission. This approach allows us to identify the specific drivers of energy consumption and provide an empirical basis for choosing configurations that maximize a sensor node's battery lifetime for a given application.}

This work is structured in seven sections. 
Section \ref{sec:background} reviews related research and states the contributions of this work. Derived from this, we explain the design of our experiments in Section \ref{sec:approach}. Section \ref{sec:methodology} describes the methodology applied to conduct the experiments, followed by Section \ref{sec:results-discussion}, which presents and discusses the results. Section \ref{sec:limitations_outlook} states the limitations and provides an outlook on potential extensions. Section \ref{sec:conclusion} concludes the work.

\section{Background}
\label{sec:background}
Research on systems\textquotesingle{} energy consumption for applications using
\ac{iot} and \ac{ml} is an evolving field, as rapid advances in hardware and algorithms require repeated evaluation. In the following, we review energy measurement in \ac{iot} nodes and related work, identify the research gap, and highlight our contribution.

\subsection{Related Work}
Due to the limited power and computing capacities of \acp{mcu}, significant research has focused on these devices\textquotesingle{} energy efficiency~\cite{callebaut_art_2021}, particularly by evaluating and optimizing sleep modes and reducing active time~\cite{wu2015energyefficientsleep}.
A wide spectrum of connectivity options has been evaluated to assess and mitigate transmission energy. This includes application-layer message protocols like CoAP and MQTT~\cite{Pavelic_COmpareCoAPMQTT} and physical networks such as BLE~\cite{Nair_iotbleopt}. For longer distances, established \acp{lpwan} such as LTE-M~\cite{labdaoui2023energy}, \ac{nbiot}~\cite{labdaoui2023energy,Sinha_LPWANBIOT, hasanModelingPrototypingIoTbased2025,mekki2019comparative}, and LoRaWAN~\cite{Sinha_LPWANBIOT, hasanModelingPrototypingIoTbased2025,mekki2019comparative} are widely deployed. Beyond these, emerging standards such as 5G NR RedCap~\cite{3gpptr5GRedCap} and Wi-Fi HaLow~\cite{wifihalow} offer promising capabilities in terms of throughput, while BLE Long Range~\cite{bleLongRange} demonstrates remarkable energy efficiency at intermediate distances. Despite these advances, the fundamental trade-off remains, as transmitting large data payloads consumes significant energy, especially over long distances. This prompts the need to explore local alternatives, such as on-device processing.

Early work on the energy consumption of on‑device processing focused largely on conventional non-\ac{ml} tasks, which are computationally lighter than modern \ac{EmML} problems. For example, Imran et al.~\cite{imran_implementation_2012} implemented different vision tasks on various hardware platforms and later compared their findings with a theoretical model of energy consumption \cite{shahzad_comparative_2014}. However, these works fail to account for the computational complexity of \ac{ml} tasks. More recent studies investigate the factors that influence ML models\textquotesingle{} energy consumption, such as the use of different frameworks~\cite{pena2017benchmarking, zhang_pcamp_2018} or hardware selection~\cite{pena2017benchmarking, zhang_pcamp_2018, liang_ai_2020}. However, energy consumption is often treated as a secondary metric or estimated using floating point operations (FLOPs). While FLOPs indicate computational complexity, they do not fully capture actual hardware energy consumption. 

Moreover, specialized models and engines have been designed to run more efficiently on small devices \cite{linMCUNetTinyDeep2020a} or dynamically adapt model complexity based on available energy \cite{leeNeuroZEROZeroenergyNeural2019}, alongside complete ecosystems for compressing and executing \ac{ml} models~\cite{gobieskiIntelligenceEdgeInference2019}.
While these works demonstrate significant advances in embedded model efficiency, they typically focus solely on inference energy and neglect the remaining tasks in \ac{iot} pipelines. Furthermore, when addressing energy consumption in other \ac{iot} tasks, such as transmission energy, these studies rely on literature values and overlook the need for real-world measurements to ensure transferability.

Recent studies have attempted to close this gap but have not covered the described case of remote monitoring with \ac{SoC} devices. Instead, they investigated more sophisticated hardware and networks, e.g., the \texttt{Jetson TK1} and LTE, and explored the optimal layer to split deep neural network computation \cite{kangNeurosurgeonCollaborativeIntelligence2017}. Recent hardware advancements enabled \ac{cnn} execution on significantly smaller devices. Muhoza et al.~\cite{muhoza_power_2023}, for example, evaluated the energy consumption differences when using BLE to transmit either human activity sensor data or the outputs of a locally run \ac{cnn} to recognize human activity.
While personal networks are not suitable for long-range applications like remote monitoring, the reported findings showed a 21\% reduction when running the \ac{cnn} locally and transmitting the result. Even smaller devices have been investigated in the field of batteryless device research, combining \ac{lpwan} and image-based sensing \cite{desaiCamaropteraLongrangeImage2022,gobieskiIntelligenceEdgeInference2019,gadreAdaptingLoRaGround2024,giordanoBatteryFreeLongRangeWireless2020,hasanModelingPrototypingIoTbased2025}. 
Although these devices demonstrate extreme cases of energy efficiency, they are severely limited in data size, limiting image resolution to QQVGA (160 x 120 pixels) or below and model sizes to only a few layers.
Similarly, \ac{tinyml} has been applied in audio processing applications, such as transmitting sounds from bees~\cite{desimoneAdvancingBeekeepingIoT2024}, birds \cite{TinyChirp}, mosquitoes \cite{tinymlMosquito}, and cicadas~\cite{cicadafilterVNNdetection} and in applications with other high-resolution data requirements (e.g., EEG \cite{sleepstage} and ECG \cite{Arrhythmia}). Most of these demonstrated optimized energy efficiency, however, they often rely on fine-tuned optimization or custom data protocols to minimize the energy required to process data, and none compare their measured energy consumption to merely sending the unprocessed data.

In summary, most prior studies have explored individual components of a system, without providing a holistic evaluation or transferable results, as it is often non-trivial to extrapolate findings to a full-system implementation. While some works analyze near end-to-end applications, they focus only on extreme cases, either relying on relatively powerful devices or heavily constrained devices that require custom compression techniques and limit input sizes to very low-resolution data. Less attention has been given to \acp{mcu}, which offer reasonable trade-offs between memory resources, energy efficiency, and long-range connectivity for many remotely operating \ac{iot} applications. They can capture data and handle well-known \ac{tinyml} models on-device, covering a broad set of applications with high-resolution data requirements, such as those processing image, audio, or movement data.

\subsection{Contribution}
\begin{figure}[h]
    \centering
    \includegraphics[width=1\linewidth]{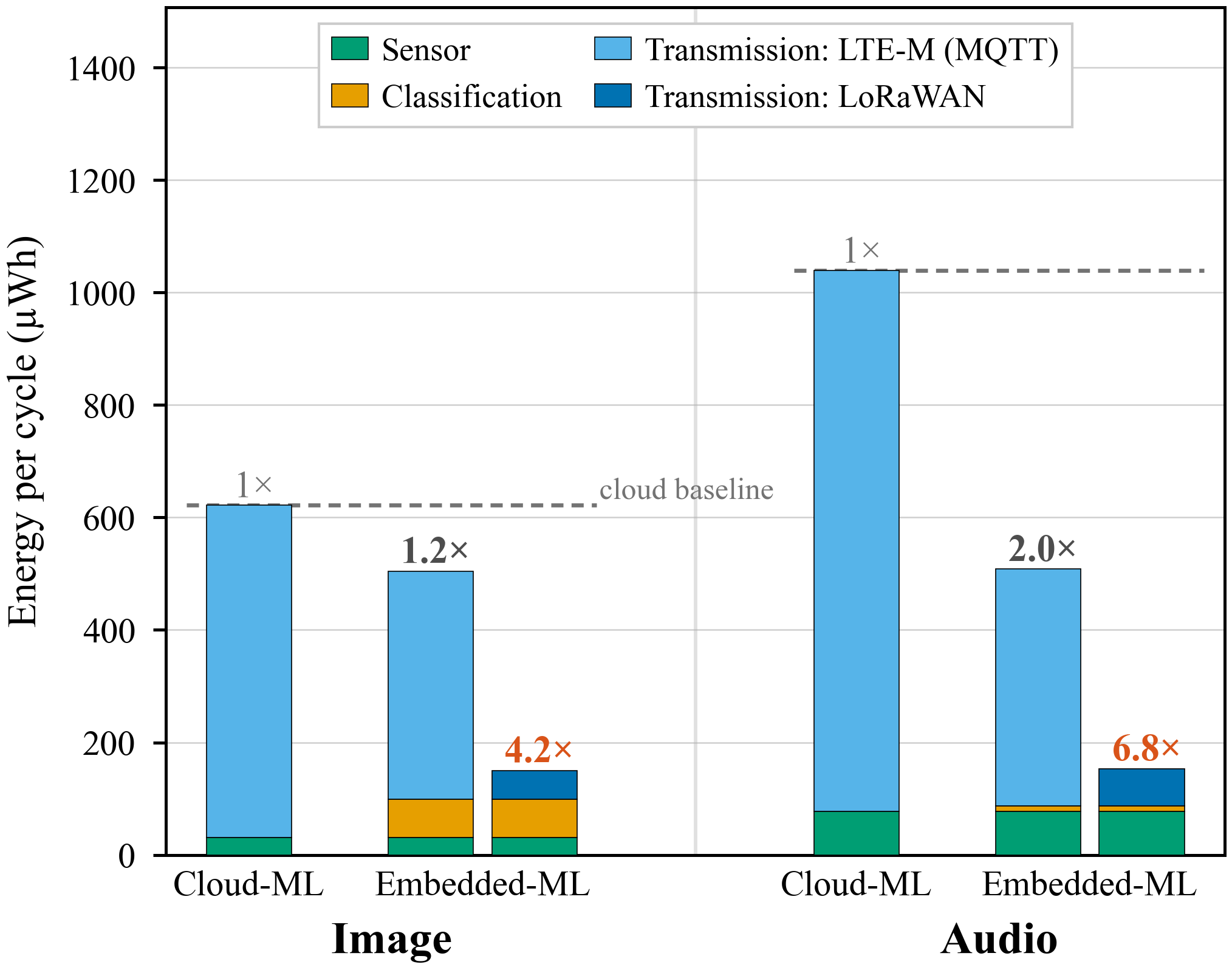}
    \caption{\small Energy consumption of an \ac{iot} node per sensing cycle. Transmitting raw sensor data via LTE-M with MQTT (Cloud-ML) dominates the budget. Running CNN classification on-device, followed by transmitting the classification result (Embedded-ML), reduces the energy by the factor given above each bar. For image classification, an OV2640 camera module, SqueezeNet, and the \acs{plantvillage} dataset are used; for audio classification, a digital MEMS microphone, \acs{tcresnet}, and the Yellowhammer dataset (see Section \ref{sec:cnns}) are used. Because LoRaWAN supports only small payloads, we could not measure the transmission of image and audio sensor data}
    \label{fig:contribution_overview}
\end{figure}

Our work contributes a systematic analysis of sensor node energy consumption, evaluating practical deployments across multiple model architectures, network protocols, and application scenarios to guide efficient use of \ac{EmML} in remotely operating \ac{iot} applications.
Specifically, we quantify the difference between transmitting raw data and processing it on the device, demonstrating that TinyML models can greatly reduce a sensor node's energy footprint compared to conventional cloud-based processing. Moreover, the reduction in payload size enables the use of unlicensed \acp{lpwan}, such as \ac{lorawan}, thereby extending the potential communication range and autonomy of applications. Figure~\ref{fig:contribution_overview} previews our results, displaying the differences in energy consumption across selected configurations.
Our empirical approach avoids the pitfalls of relying solely on simulation or aggregating fragmented results from disparate sources, providing a quantified real-world approach for \ac{cnn}-based data processing and network operations.
All software and firmware for realizing these scenarios are publicly available\footnote{https://anonymous.4open.science/r/SendLessSaveMore}.

To sum up, the main contributions are as follows:
\begin{enumerate}
    \item We systematically analyze the energy consumption of \ac{iot} image and audio processing applications with and without an embedded CNN model (i.e., transmitting inference results vs. raw data).
    \item We created a reusable prototype and pipeline to profile the energy requirements of \ac{iot} applications for image and audio processing and transmission over long distances.
    \item We offer insights into the energy consumption of different sub-tasks of an end-to-end IoT deployment using several widely used network protocols and \ac{ml} models, enabling recommendations for different settings.
\end{enumerate}

\section{Approach}
\label{sec:approach}
We structure our evaluation as a systematic, task-decomposed set of experiments and measurements, ranging from individual task energy measurement to end-to-end battery-lifetime estimation, enabling generalizable conclusions rather than a single system comparison.
Here, we first describe the two underlying paradigms we distinguish and then present our experimental design to analyze them.

\subsection{Compared Paradigms}
\begin{figure}[h]
    \centering
    \includegraphics[width=0.9\linewidth]{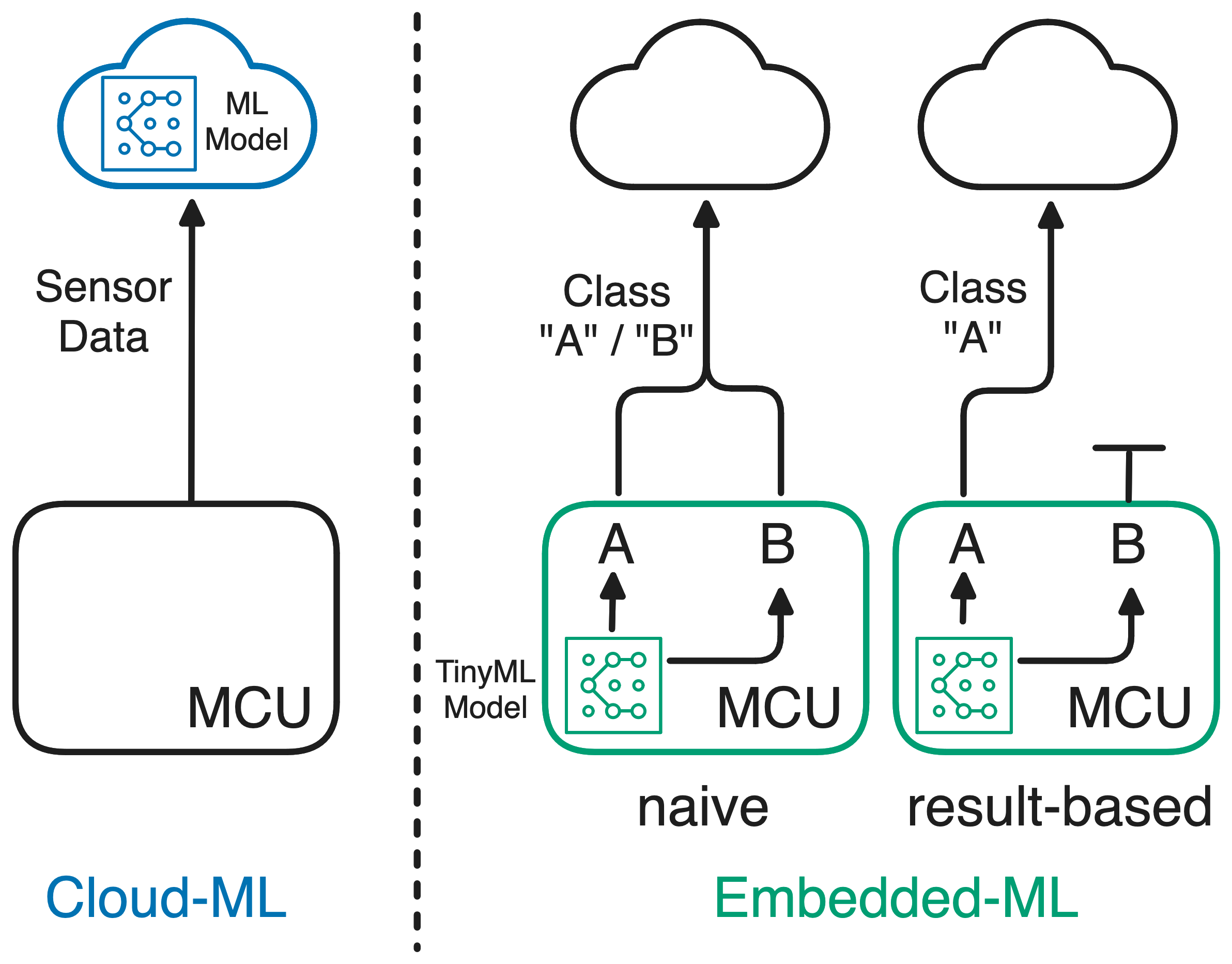}
    \caption{\small Two paradigms for \ac{ml}-based processing in the \ac{iot}. Devices either send sensor data to the cloud for machine learning inference (\texttt{Cloud-ML}) or process the data locally with a \ac{tinyml} model (\texttt{Embedded-ML}). In the latter scenario, the device can either send all data (\texttt{naive}) or decide based on the model output whether data are transmitted (here "A") or suppressed ("B") (\texttt{result-based}).}
    \label{fig:paradigms}
\end{figure}

We differentiate between two deployment paradigms for a sensor node: \texttt{Cloud-ML} and \texttt{Embedded-ML} (see Figure \ref{fig:paradigms}). The conventional \texttt{Cloud-ML} approach, previously described as cloud-based processing, entails collecting data through a sensor connected to an \ac{mcu}, then transmitting all data to a remote server via an appropriate network protocol. The data are assumed to be processed further on the server. Data transmission is a main driver of the sensor node's energy cost for this approach.
Conversely, in the \texttt{Embedded-ML} approach, the data are processed with a \ac{tinyml} model executed directly on the MCU. This introduces a new energy cost factor, i.e., on-device inference. After sensing data and running inference, there are two possible modes of data transmission: \texttt{naive} sending and \texttt{result-based} sending. Naive sending transmits results for each application cycle, whereas result-based sending transmits only actionable results. To illustrate this distinction, consider the application presented in Figure \ref{fig:exampleiota}, where plants are classified as having a certain disease or being healthy. It is sufficient to transmit data only when a plant is classified as infected; it is not necessary to report on every application cycle in which all plants are classified as healthy. The frequency of events of interest is heavily application-specific, but it is common for them to be rather rare in most \ac{iot} applications for remote monitoring.

\subsection{Experimental Design}

\begin{figure}[h]
\centering
    \includegraphics[width=1\linewidth]{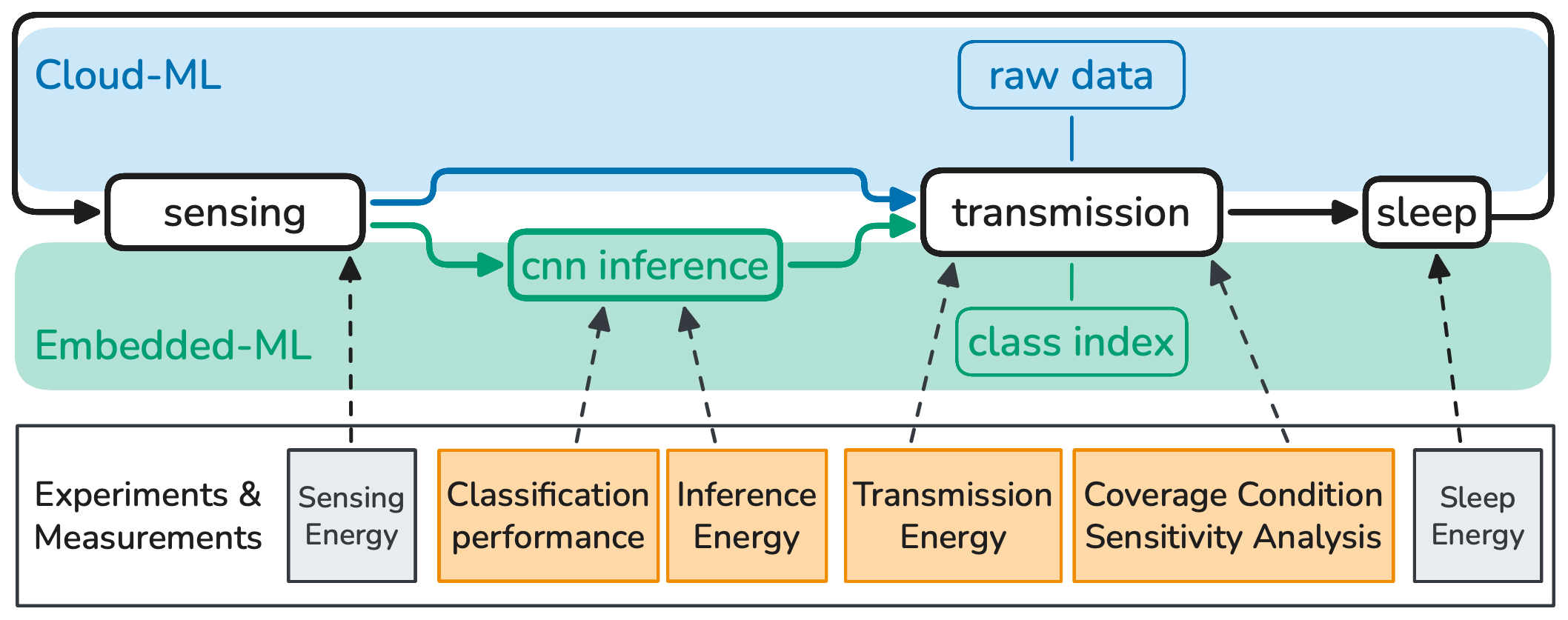}
\caption{\small Overview of the experimental design. Experiments and measurements report per-task metrics. Grey boxes are measurements; orange boxes are experiments where a set of independent variables is systematically varied.}
\label{fig:experiment-design}
\end{figure}

To compare the two paradigms systematically, we decompose the approaches into their tasks \textit{sensing}, \textit{inference}, \textit{transmission}, and \textit{sleep} and structure our evaluation as a set of experiments and measurements at the task level (see Figure~\ref{fig:experiment-design}). We distinguish between direct measurements, where energy consumption is recorded without parameter variation, and experiments, where we systematically vary key variables to test their impact on energy consumption and inference performance of the respective tasks.

Sensor readings are required in all scenarios, independently of the chosen paradigm. Therefore, we directly measured the energy required to collect data in the sensing task from two widely used low-cost sensors (camera and microphone) and report average values.
The inference task on the \ac{iot} device is only present within the \texttt{Embedded-ML} paradigm; we neglect inference energy measurement in the cloud for \texttt{Cloud-ML}, as this work focuses on energy consumption at the sensor node. We prepare several \acp{cnn} with different architectures, data types, and datasets for use in a sensor node and conduct two experiments to evaluate, for each configuration, the classification performance and measure energy consumption at the sub-task level of the inference pipeline. 
To cover the data transmission task, which is present in both paradigms, we conduct an experiment to measure energy at the sub-task level of the transmission process, while systematically varying data sizes, radio technologies, data protocols, and power-saving modes. Additionally, we perform a sensitivity analysis regarding coverage conditions to generalize our findings on transmission energy.
Finally, we compose the measured task-based energy and supplementary sleep-task measurements to provide a holistic picture of application-level energy. This allows for direct comparison of both paradigms as well as the naive and result-based sending modes. We also estimate the battery lifetime of a sensor node for different sleep-period lengths and configurations to inform practical \ac{iot} deployments with different duty-cycle requirements. This bottom-up structure allows us to attribute any energy difference between the paradigms back to specific tasks and design decisions, rather than reporting only aggregated outcomes.

\section{Methodology}
\label{sec:methodology}
Herein, we detail the implementation of the experimental design introduced above, covering the prerequisites and design decisions, the machine learning methods, the energy-measurement methodology, and the specific parameter configurations and reported metrics of each experiment.
\subsection{Prerequisites}
\label{subsec:preliminary}

\noindent
\subsubsection{Communication}
To make a fair comparison, we select the radio technology and communication protocol that best suit each scenario. All scenarios involve transmitting data from remote locations to a server while minimizing energy consumption. Consequently, the use of \ac{lpwan} is inevitable. We consider 5G NR RedCap, WiFi-HaLow and BLE Long Range out of scope for the described application as they are either designed for systems with higher power profiles and throughput needs (5G NR RedCap, WiFi-HaLow), limited in practical range (BLE Long Range), or lack the commercial maturity and infrastructure required for widespread environmental deployment.
Based on this prerequisite, we considered three \ac{lpwan} technologies: \ac{nbiot} (LTE CAT-NB), LTE-M (LTE CAT-M1), and LoRaWAN. 
The first two technologies operate in licensed spectrum and use a subset of LTE standards.
\ac{nbiot} constrains the bandwidth to a single narrow band of 180 kHz transmission bandwidth with an upload rate of up to 160 Kbps (CAT-NB2). LTE-M allows for a larger transmission bandwidth of up to 1.08 MHz with an upload rate of up to 1.1 Mbps. Both technologies use existing mobile network operator infrastructure. However, this introduces data-transmission costs, as a subscription is necessary and usage-based charges are common.

In contrast, LoRaWAN operates in the unlicensed spectrum, communicating over ISM radio bands, building on the physical LoRa layer \cite{LoRaWANSpec}. This lets users decide whether to build their own infrastructure or join available networks. 
LoRaWAN communication range, payload size restrictions, and available bandwidth depend on the application region and the data rate used. 
Data rates can be altered by adjusting the \ac{sf} parameter. Increasing \ac{sf} extends the communication range, albeit at the expense of bit rate, maximum payload size, and power consumption. At \ac{sf}7, the maximum bit rate is 5.5 Kbps and the maximum message size is 222 bytes, whereas at \ac{sf}12, these figures are 250 bps and 51 bytes, respectively. In addition, \ac{lorawan} enforces duty cycles for sending data, reducing the overall transmission time to less than 1\%; thus, it is not appropriate to send high-resolution sensor data such as images or audio samples. However, sending small-sized messages such as \ac{ml} inference results is reasonable if those are not required at high-frequency intervals \cite[p. 8,19,24,39]{LoRaWANRegionalSpec}.

Appropriate application-layer protocol selection depends on the nature of the data to be transmitted. MQTT and CoAP (Constrained Application Protocol) are widely used application-layer protocols in conjunction with either LTE-M or \ac{nbiot}. MQTT operates over TCP/IP or other network protocols that provide ordered, lossless, bidirectional connections \cite{mqtt2019}, and it allows sending messages up to 256 MB. Therefore, it is well suited for transmitting both small images and messages.
CoAP transfers data with UDP/IP. Because it is lightweight and connectionless, it is well-suited to operate on low-power wide-area networks. The initial specification was documented in RFC 7252 \cite{shelby_constrained_2014}. Originally, the message size was restricted to 1152 bytes for the payload. For larger messages, block-wise transfers have been introduced \cite{blockwisetransfer}. \ac{lorawan} applications usually rely on a \ac{lorawan} network server to handle the data upon arrival, as the available message size and duty-cycle restrictions render most application-layer protocols inappropriate. We avoid custom protocols to ensure our insights are applicable and transferable for embedded developers.

\subsubsection{Hardware}
\begin{figure}[h]
    \centering
    \includegraphics[width=1\linewidth]{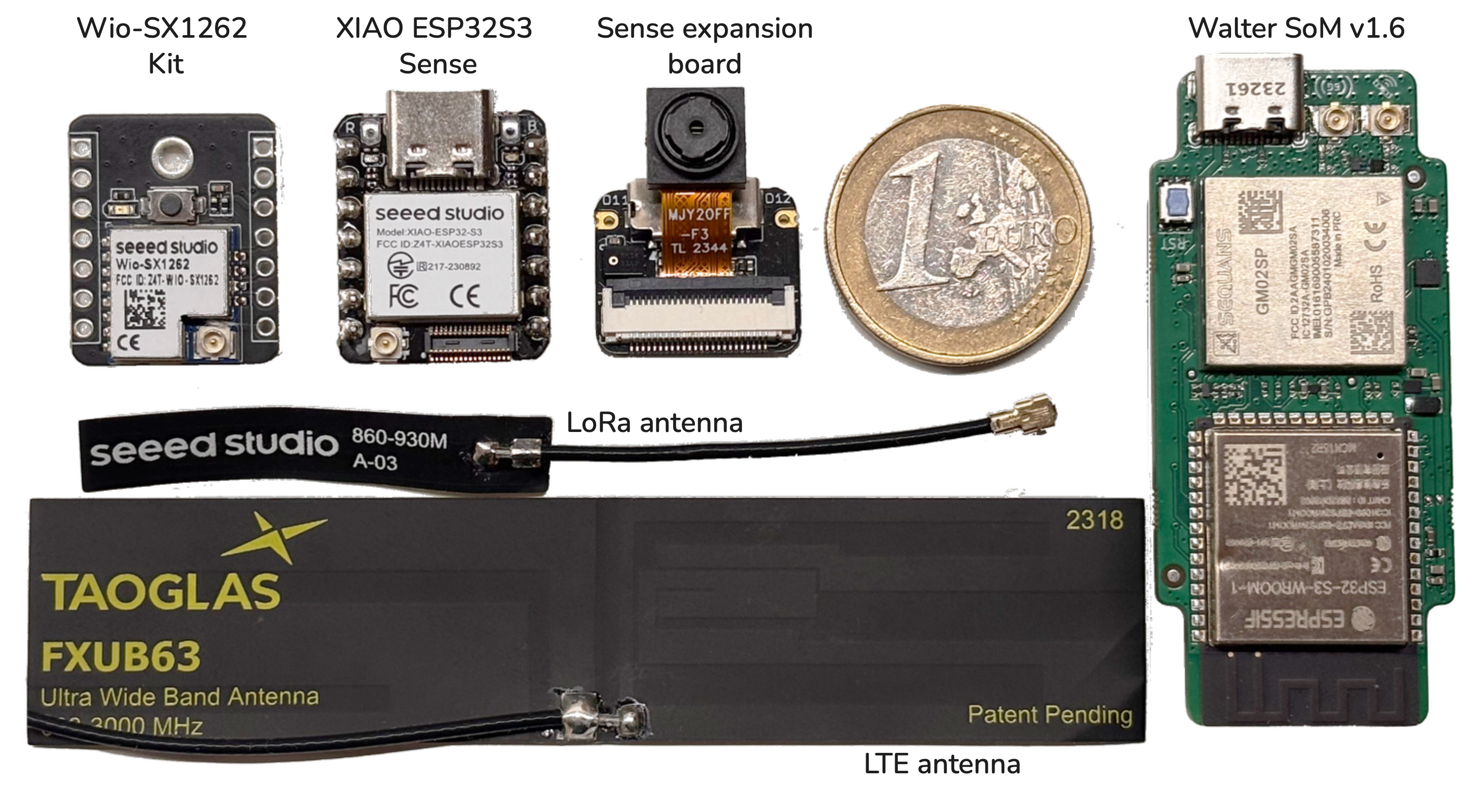}
    \caption{\small Selected hardware for the case study: Seeed XIAO ESP32-S3 Sense with two interchangeable expansion boards, the Wio-SX1262 Kit with LoRa antenna and Sense expansion board, and Walter SoM v1.6 with Taoglas LTE antenna. One~euro for scale.}
    \label{fig:mcus-size}
\end{figure}

The setup requires that the \ac{mcu} can capture and store data of reasonable size (e.g., image or audio), as well as store and run \acp{cnn}. It also needs communication modules to send the data or \ac{cnn} outputs over the previously discussed communication protocols. We chose the ESP32-S3 series because it is available with a broad range of memory configurations and is accessible to developers on various integrated microcontroller boards with compatible communication modules.
It also provides a set of low-power sleep-mode configurations and a well-documented code stack and community for developing \ac{tinyml} projects. Since an integrated microcontroller board fulfilling all requirements at once was not available, we decided to use two open-source development kits, based on microcontrollers of the ESP32-S3 series: (1) the \texttt{Walter SoM v1.6}\footnote{https://www.quickspot.io/index.html} and (2) the \texttt{Seeed XIAO ESP32-S3 Sense}\footnote{https://www.seeedstudio.com/XIAO-ESP32S3-Sense-p-5639.html} with an additional \texttt{Wio-SX1262 Kit}\footnote{https://www.seeedstudio.com/Wio-SX1262-with-XIAO-ESP32S3-p-5982.html} as an interchangeable extension (see Figure \ref{fig:mcus-size}). 

The boards differ in their \ac{mcu} chip version, available PSRAM memory, peripherals, and communication technologies. Table \ref{tab:esp-series-comparsion} summarizes the relevant differences for the present work.

\begin{table}[h]
\centering
\caption{Microcontroller specifications}
{\scriptsize
\begin{tabular}{p{44pt}|p{90pt}p{60pt}}
 & {\centering\textbf{Seeed XIAO ESP32-S3 Sense \& Wio-SX1262 Kit}\par} & {\centering\textbf{Walter SoM v1.6}\par} \\ \toprule
\textbf{MCU} & ESP32-S3R8 & ESP32-S3R2\\ \midrule
\textbf{Memory} & 512KB + 8MB & 512KB + 2MB\\ \midrule
\textbf{Peripherals} & Camera, Mic, SX1262 LoRa Radio & Sequans GM02SP Modem \\ \midrule
\textbf{LTE-M} & $\times$ & \checkmark \\ \midrule
\textbf{\ac{nbiot}} & $\times$ & \checkmark \\ \midrule
\textbf{LoRaWAN} & \checkmark  & $\times$ \\ \midrule
\textbf{Experiment Usage} & {\centering\ac{cnn} inference \& LoRaWAN energy\par} & {\centering LTE-M \& \ac{nbiot} energy\par} \\ \bottomrule
\end{tabular}}

\label{tab:esp-series-comparsion}
\end{table}

The Walter board combines the ESP32-S3 with a Sequans GM02SP LTE-M/\ac{nbiot} 5G modem. The modem supports power-saving schemes such as \ac{PSM} and \ac{eDRX}. These can minimize power consumption during periods of inactivity while avoiding expensive reattachment to the network. In this work, we use the board to test the energy consumption of sending messages with LTE-M or \ac{nbiot}. Due to the board's reduced memory capacity, support for \ac{tinyml} is limited in our case study.
However, the XIAO board has 512 KB of SRAM and 8 MB of on-chip PSRAM, and an extension board equipped with an OV2640 camera and a digital MEMS microphone, making it a realistic test bed for the sensing and \ac{tinyml} applications at hand. The camera can output JPEG; therefore, we did not apply further image compression options. To ensure comparability, we tested all \ac{ml} models on the Seeed XIAO board. Exchanging the camera and microphone extension for a compatible Wio-SX1262 Kit enabled the board to communicate via LoRaWAN and allowed us to measure the energy consumption of corresponding messages.
To generalize our findings for different network coverage conditions, we additionally used a PE4302 Digital Step RF Attenuator for the sensitivity analysis. This allowed us to artificially weaken the signal.

\subsubsection{Software}
For data processing and model training, we opted for \texttt{PyTorch}\footnote{https://pytorch.org} due to widespread adoption in \ac{cnn} training. We implemented the firmware for the use cases using the Espressif \ac{SoC} \texttt{ESP-IDF} development framework and the Arduino IDE (only for LoRaWAN communication).
The ESP-IDF framework allows specifying details such as explicitly assigned memory spaces on the partition and custom deep-sleep wake-up calls. This is particularly important when implementing \texttt{Embedded-ML} tasks, where multiple factors (multicore usage, CPU clock frequency) need to be considered to reduce energy consumption. \texttt{esp-dl}\footnote{https://components.espressif.com/components/espressif/esp-dl v3.1.0} is a widely used library for ESP \acp{mcu}, known for its efficiency. This library is supplemented by a quantization tool, \texttt{esp-ppq}\footnote{https://github.com/espressif/esp-ppq}, which handles inputs and models from \texttt{PyTorch}.
Further, using libraries from the hardware manufacturer ensures the functionality is tailored to the hardware; this would also be the most straightforward decision in practice. However, this decision limits the approach's generalizability to the ESP product family. While alternative solutions, such as LiteRT from TensorFlow, offer greater interoperability across different hardware architectures, achieving cross-platform compatibility almost always comes at the cost of suboptimal utilization of a specific hardware platform. For this reason, we opted to use the available open-source software provided by the ESP vendor.
Despite this, we used the Arduino IDE for LoRaWAN communication, since the task did not require extensive computation and the Arduino IDE already had an existing library stack for handling LoRaWAN, whereas ESP-IDF did not.

\subsubsection{Use Cases \& Datasets} \label{subsec:datasets}
As representative problems, we consider the environmental monitoring use cases depicted in Figure \ref{fig:exampleiot} and use widely available image and audio datasets. For the image-based applications, we selected the \textit{\acs{plantvillage}} dataset~\cite{hughes_open_2016, ferentinos_deep_2018} in the version by Pandian et al. \cite{j_data_2019} and the \textit{\ac{cub}} dataset~\cite{WahCUB_200_2011}.
The \acs{plantvillage} dataset contains images of individual plant leaves from 14 different species with a total of 26 diseases, images of healthy leaves for 12 species, and a class of background images without leaves. This is one of the most frequently used public datasets for plant disease detection-related tasks~\cite{boulent_convolutional_2019, pandey_survey_2024}. 
The \ac{cub} dataset comprises 11,788 images of birds belonging to 200 different species. The larger number of output classes makes the underlying classification task more challenging, especially given the previously described runtime's computational constraints.
For audio classification, we selected the \textit{\ac{yellowhammer}} dataset~\cite{yellowhammer_dataset_2025}.
It contains over two hours of \ac{yellowhammer} songs and negative samples, including other bird species and background noise. Each sample is two seconds long and was recorded in different environments at different distances.

The use cases allow flexibility in duty-cycle requirements, i.e., the frequency of sensor readings and the resulting length of a sensor node's sleep period.  For plant disease detection, it may be sufficient to capture and process images once or twice a day. However, a system classifying birds at a bird feeder may use an external trigger such as a scale (e.g., in \cite{niers_birdf_2022}) to wake a sensor node from sleep and trigger a new image, resulting in fluctuating sleep-period lengths. In fact, an audio-based bird classification device may need to operate at a very high frequency to avoid missing a bird song, but may only need to operate during certain periods each hour.

\subsection{CNNs}
\label{sec:cnns}
\begin{figure}[t]
    \centering
    \includegraphics[width=1\linewidth]{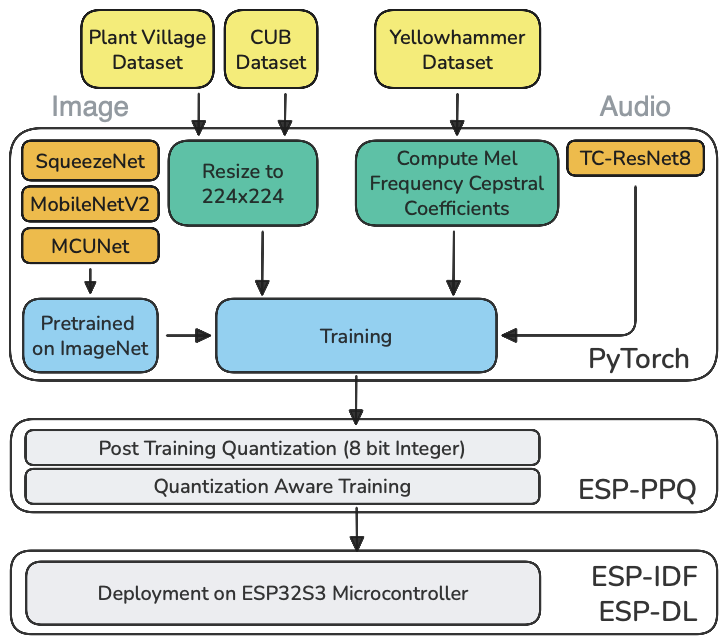}
    \caption{Overview of the model preparation pipeline for Embedded-\ac{ml} deployment.}
    \label{fig:model_prepare}
\end{figure}

We evaluated multiple \ac{cnn} architectures for their applicability to the given context. Figure~\ref{fig:model_prepare} provides an overview of the software tools and model architectures employed, linking them to the stages of the training process. 

\subsubsection{Model Architectures}
For the image classification tasks, we choose \acs{mobilenet} \cite{sandler_mobilenetv2_2018}, \acs{squeezenet} \cite{SqueezeNet} and \acs{mcunet} \cite{linMCUNetTinyDeep2020a}, all of which are well-established, small-sized models.
For sound classification, we selected a TC-ResNet8 \cite{choi19_interspeech}.
All of them are known for providing good performance while being relatively small. In fact, \texttt{esp-dl} officially supports a version of \acs{mobilenet} trained on ImageNet data \cite{ImageNet}. 

During our evaluation, we considered DenseNet~\cite{DenseNet}, but excluded it because we could not reduce it enough, with the chosen compression methods, to fit in the available memory.
ShuffleNet~\cite{ShuffleNet}, MNASNet~\cite{MNASNet}, and MobileNetV3~\cite{howard2019searching} were also excluded, as some of their layers lacked support to be run with the chosen \ac{mcu} software stack. We avoided customizing the models by replacing non-supported layers, as this would have biased the comparison to state-of-the-art versions of the models. Notably, once MNASNet is supported, we expect even better results in both energy efficiency and accuracy, given its demonstrated advantages in terms of accuracy and inference latency compared to MobileNetV2~\cite{MNASNet}. 

\subsubsection{Training}
\label{subsubsec:transfer}
The default weights of the three image classification models \acs{mobilenet}, \acs{squeezenet}, and \acs{mcunet} were based on previous training on the ImageNet dataset~\cite{ImageNet}.
The images from the datasets were resized to 224 $\times$ 224 pixels, slightly smaller than QVGA resolution, as the models were pretrained with this input size. This resolution is appropriate for feature extraction in the proposed use cases while remaining realistic for capturing and processing image data on an \ac{mcu}~\cite{hasanModelingPrototypingIoTbased2025}.
For the pretrained models, we used a two-step transfer learning approach to train them on the given datasets. First, only the classification head was retrained, while the weights of the other layers were frozen.  
Subsequently, we retrained additional layers to fine-tune the models.
The audio model was trained from scratch, as no pretrained weights were available from similar, larger datasets. The model uses \ac{mfcc} as input.

For all image classification models, a 70/20/10 training/validation/test split was used.
The audio dataset was provided with a predefined split, based on individuals, resulting in a ratio of approximately 60/20/20, with two individuals reserved for the evaluation set. During training, we tuned hyperparameters to determine effective learning rates and weight decay values. 
All values are listed in Table \ref{tab:params}. 
All models were trained with at least 200 different hyperparameter combinations for 100 epochs of transfer learning and 100 epochs of fine-tuning. Early stopping was enabled to stop training after 5 epochs with no improvement in model performance. 
The hyperparameter set yielding the highest accuracy was selected, and the corresponding model was passed to the quantization step.

\begin{table}[htbp]

    \setlength\extrarowheight{2.3pt}
    \centering
    \caption{Hyperparameters used during training.}
        {\scriptsize
    \begin{tabular}{l|l|l|c}
        \multicolumn{3}{l|}{Parameter} & Values \\ \toprule
        \multicolumn{3}{l|}{Optimizer} & AdamW \\ \midrule
        \multicolumn{3}{l|}{Learning rate scheduling} & Multi‑step method \\ \midrule

        \multirow{6}{*}{\shortstack[l]{Image\\Classification}}
        & \multicolumn{2}{l|}{Batch size} & $64$ \\ \cline{2-4}
        & \multirow{2}{*}{\shortstack[l]{Transfer\\learning}} & Learning rates & $1\times10^{-4},\,1\times10^{-2}$ \\
        &                                            & weight\_decay    & $1\times10^{-4},\,1\times10^{-2}$ \\ \cline{2-4}
        & \multirow{3}{*}{\shortstack[l]{Fine\\tuning}}   & Learning rates & $1\times10^{-5},\,1\times10^{-3}$ \\
        &                                            & weight\_decay    & $1\times10^{-5},\,1\times10^{-3}$ \\
        &                                            & Number of layers & $2\text{–}10$ \\ \midrule

        \multirow{6}{*}{\shortstack[l]{Audio\\Classification}}
        & \multicolumn{2}{l|}{Batch size} & $256$ \\ \cline{2-4}
        & \multirow{2}{*}{\shortstack[l]{Transfer\\learning}} & Learning rates & $1\times10^{-4},\,1\times10^{-1}$ \\
        &                                            & weight\_decay    & $1\times10^{-4},\,1\times10^{-2}$ \\ \cline{2-4}
        & \multirow{3}{*}{\shortstack[l]{Fine\\tuning}}   & Learning rates & $1\times10^{-5},\,1\times10^{-2}$ \\
        &                                            & weight\_decay    & $1\times10^{-4},\,1\times10^{-2}$ \\
        &                                            & Number of layers & $3$ \\ 
        \bottomrule
    \end{tabular}}
    \label{tab:params}
\end{table}

For audio preprocessing, we used the Kaldi \ac{mfcc} implementation in PyTorch~\cite{Povey2011TheKS, ravanelli2019pytorch} since a corresponding C++ implementation in esp-dl exists to embed the code on the~\ac{mcu}.

\subsubsection{Quantization}
\label{subsubsec:quant}
All models were quantized from 32‑bit floating-point values to 8‑bit integers to avoid expensive floating-point operations on the \ac{mcu}. We use the {\tt esp‑ppq} package, which supports both \ac{ptq} and \ac{qat}. For the initial \ac{ptq}, we chose a layerwise equalization scheme as proposed in \cite{LayerEqualPPQ}, since per‑channel quantization yields lower quantization errors for small models than per‑tensor quantization. Manual exploration of the equalization hyperparameters led to a multi-branch configuration with four iterations and a value threshold of 0.4 as the best-performing setting for all evaluated models.  
After \ac{ptq}, we perform a single epoch of \ac{qat} on the original training dataset to recover from the accuracy loss introduced by quantization.
The complexity of fine‑tuning and hyperparameter optimization can be scaled arbitrarily. In the present study, we allocated a substantial yet pragmatic amount of computational effort to obtain a realistic estimate of the attainable accuracy. While the reported accuracies demonstrate the models' capabilities, any additional improvements would likely be incremental rather than substantial.

\subsubsection{Deployment}
\label{sec:energy:deployment}
To deploy the models and perform inference on the selected hardware, we created firmware compatible with each chosen \ac{cnn} and its input data type. It allowed us to profile the energy consumption of the \ac{cnn}-based classification. We opted not to implement the sensing task in this firmware, as we measured the required energy separately. Instead, we replaced it with loading a common image or audio file from the dataset. This approach avoids influencing the runtime by varying the input and ensures that representative examples are processed.
In the firmware, after booting up the \ac{mcu}, the model is loaded from flash memory, the image or audio file is preprocessed, inference is performed, and the model output is postprocessed using a softmax layer to generate a class label. Before and after each phase, markers for profiling energy consumption are implemented. 
During energy profiling, the firmware was deployed on the Seeed XIAO board without the sensor extension board being connected.

\subsection{Transmission}
We evaluated multiple communication protocols to measure the energy consumption of wireless data transmission. Below, we describe the testing setup, including the network operators, endpoints, and device firmware. We then present the independent variables varied across experiments and a sensitivity analysis on coverage conditions.

\subsubsection{Network Operators \& Endpoints}
Cellular network operators were chosen based on availability, support for IoT, and the best allowed configurations for energy efficiency. We limited the LTE frequency band selection to Band 3, which was available for all operators. We observed different RRC inactivity timer durations across operators. The best option for \acs{nbiot} was Vodafone Germany with Public Land Mobile Network (PLMN) ID 26202, allowing the device to release after about two seconds, whereas the shortest release time for \acs{ltem} was about 10 seconds with Telekom Germany (PLMN ID 26201).
At the recording location, without the RF attenuator connected, we recorded an LTE-M \ac{rsrp} of about $-72$ dBm and a \ac{rsrq} of $-8.3$ dB. The \ac{nbiot} network showed \ac{rsrp} values around $-79$ dBm and \ac{rsrq} values around $-8$ dB.

To allow for a realistic testing setup and validate wireless data transmission, we used external testing endpoints. We opted for widely adopted, commercially available products wherever possible, to reflect real-world deployment scenarios. For MQTT-based scenarios, we leveraged the free Cloud MQTT Broker from HiveMQ\footnote{https://www.hivemq.com/products/mqtt-cloud-broker/} to verify successful data transmission. In CoAP scenarios, we deployed an open-source Python library, \texttt{aiocoap}\footnote{https://github.com/chrysn/aiocoap}, to host a CoAP server. For LoRaWAN communication, we transmitted data to The Things Network\footnote{https://www.thethingsnetwork.org/}, a well-known LoRaWAN network server powered by The Things Stack. 

\subsubsection{Deployment}
\label{sec:transmission:deployment}
Two ESP32-S3 firmware variants were created to profile the energy consumption of data transmission: (1)~LTE-based transmission, (2)~LoRaWAN-based transmission. Both are available under an open-source license\footnote{https://anonymous.4open.science/r/SendLessSaveMore/sending}.

The LTE-based transmission firmware transmits various data types (\ac{cnn} output, images, or audio) via LTE-M or \ac{nbiot} and is deployed on the Walter board. In this case, upon first boot, the \ac{mcu} initializes the cellular modem, sets the radio technology (LTE-M or \ac{nbiot}), and configures the power-saving mode (PSM or eDRX) before connecting to the network. Once a connection is established, the device sends data (the \ac{cnn} output, the image, or an audio file), and the \ac{mcu} enters deep sleep for a predefined 30-second duration. Meanwhile, the modem remains powered on until the network releases it or the RRC Inactivity Timer runs out, and then goes to sleep mode as well. In the second to n-th cycle, the \ac{mcu} wakes up, initializes and wakes up the modem, reattaches to the network, recreates the connection to the server if necessary, and sends data. It then proceeds as in the first cycle. For the data transmission experiment, all variables are systematically varied, and each combination is run at least 10 times to mitigate variability.

The LoRaWAN-based transmission firmware behaves largely analogously to the previous firmware. 
However, it can transmit only the CNN output and uses the Seeed XIAO board with the Wio-SX1262 kit.
After sending an uplink via LoRaWAN, the device must wait for two delayed RX windows. We reduced the delay time to the minimum allowed value of one second and put the \ac{mcu} in light sleep mode while waiting between RX windows.

\subsubsection{Sensitivity Analysis}
The energy required for data transmission varies with prevailing network conditions.
To evaluate the system's robustness under poor coverage conditions, we performed a sensitivity analysis. For cellular transmission, we use a PE4302 digital step attenuator to apply attenuation levels of 0 dB, 15 dB, and 30 dB. Following the observations of~\cite{labdaoui2023energy}, which reported negligible effects at attenuations below 30~dB, we limited our study to these three attenuation levels.
Thereby, we simulated varying degrees of signal quality to observe the impact on energy consumption when sending messages of different sizes. This allows us to quantify the energy overhead caused by increased transmission latency and potential packet retransmissions, providing a more realistic estimation of battery lifetime. Given the high dimensionality of scenarios, this analysis was conducted specifically for the eDRX configuration, as we expect the variations to be largely independent of the power-saving mode (PSM vs. eDRX) used during the sleep phase.
We ran each combination of variables at least 14 times, as we observed larger fluctuations, especially at high attenuation values.
For LoRaWAN, we also performed a sensitivity analysis on varying coverage conditions, leveraging the fact that LoRaWAN allows more precise local parameterization than cellular protocols. Accordingly, we disabled the Adaptive Data Rate mechanism to maintain consistency and set the transmission power to the maximum legal value in Europe (14 dBm). We then evaluated the system's sensitivity to different Spreading Factors (\ac{sf}) of 7, 10, and 12; this effectively simulates decreasing signal conditions, as higher Spreading Factors are necessary to maintain connectivity in poor signal environments.

\subsection{Evaluation Metrics}
To provide a fair evaluation and comparison of the described paradigms and tasks, we measured several metrics across our experiments. We assessed the accuracy of the trained \acp{cnn} across our selected use cases, measured the energy consumption for various tasks, and estimated the battery lifetime for different transmission frequencies. The following section provides a detailed description of the measurement methods that were employed.

\subsubsection{Accuracy Evaluation}\label{sec:methodology-accuracy}
Model optimization and compression techniques, such as quantization, often reduce model accuracy. While lower accuracy is generally undesirable, it can be an acceptable trade-off for the benefits of model optimization, such as the ability to run a model on resource-constrained \acp{mcu}.
Therefore, one needs to cautiously evaluate whether the compressed model can still produce results with the desired accuracy. We measured the accuracy of the CNNs on the test split of the datasets before and after quantization to observe differences between the on-device model and the corresponding version running on a higher-resource platform.
As metrics, we used two variants of top-k accuracy: top-1 and top-5 accuracy. Top-k accuracy is defined as in Equation \ref{eq:top-k} where $y_i$ is the true label for the $i$-th sample, $n$ the number of samples, 
$\hat{y}_{i,k}$ is the set of top-$k$ predicted labels for the \mbox{$i$-th} sample and
$\delta(y_i, \hat{y}{i,k})$ is an indicator function that returns 1 if $y_i \in \hat{y}_{i,k}$ and 0 otherwise. Therefore, the share of samples is given where the true label is in the top-k predicted labels. 

\begin{equation}
\text{Top-k Accuracy} = \frac{\sum_{i=1}^{n} \delta(y_i, \hat{y}_{i,k})}{n}
\label{eq:top-k}
\end{equation}

\subsubsection{Energy Evaluation}
\label{sec:comparison}
\begin{figure}[h]
     \centering
     \includegraphics[width=1\linewidth]{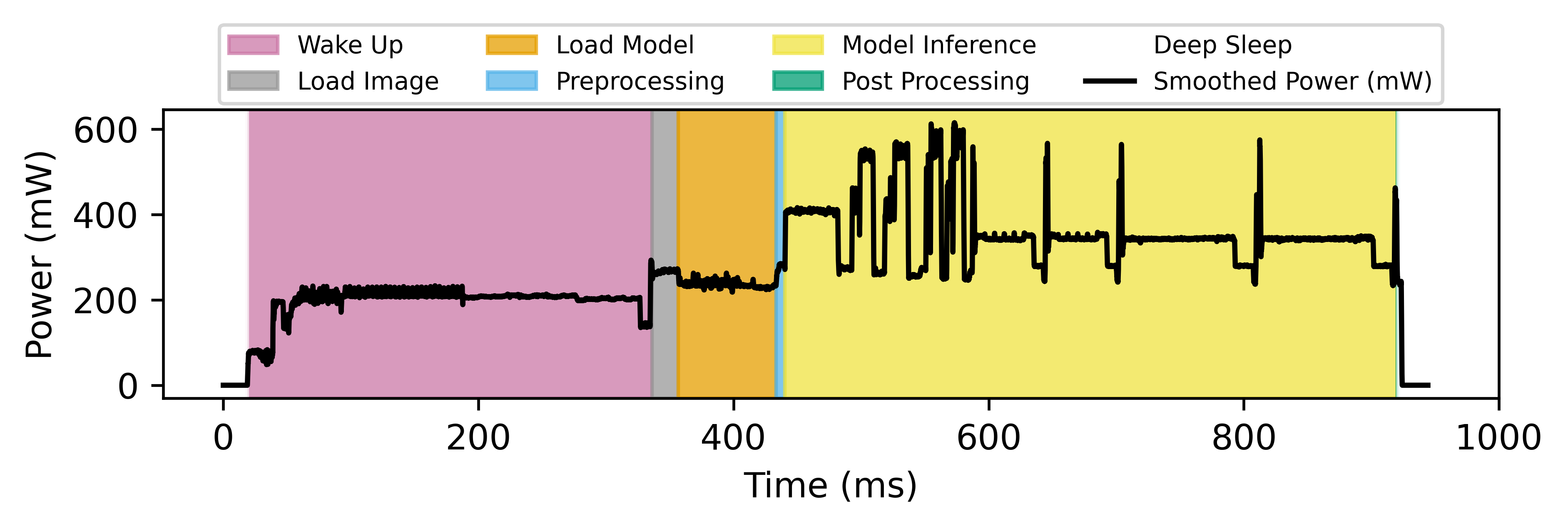}
     \caption{Depiction of the power profile measured with the Power Profiler Kit II. The selected part corresponds to one SqueezeNet \acs{plantvillage} classification cycle, colorized by different application subtasks. Measured data was smoothed using a running window of size 50 for readability.}
     \label{fig:ppk2-example}
 \end{figure}

For all power profiling tasks, we used the \texttt{Power Profiler Kit II} (PPK2).\footnote{https://www.nordicsemi.com/Products/Development-hardware/Power-Profiler-Kit-2} This device enables power supply at a static voltage level while it accurately measures power consumption at different reading resolutions ranging from $\mu$A to A. Measurements were taken at voltage supply levels between 3.3 and 3.7~V, as commonly used in lithium-ion batteries for \ac{iot} applications.
For all energy profiling, the sample rate was set to 10,000 samples per second or above, allowing for high temporal resolution in current readings. 
The PPK2 digital inputs were connected to the \ac{mcu} GPIO pins and used as a basic logic analyzer. This allows synchronizing code execution and measuring tasks individually.
An exemplary measurement is presented in Figure \ref{fig:ppk2-example}.

To mitigate variability in power consumption caused by potential measurement inaccuracy or external factors such as device temperature and wireless network load, we recorded at least 10 consecutive application cycles and report the average energy consumption per task. To minimize the effects of uncontrolled fluctuations in wireless network conditions, we conducted all experiments at the same location. 
Furthermore, we excluded the first cycle of each experiment from our analysis, as it accounts for initial setup overhead mitigated by device configuration persistence during deep sleep phases.
Based on the current readings, static supply voltage, and sample frequency, we calculated the average power, duration, and energy consumption for each task.

Note that within the \texttt{Cloud-ML} paradigm, we neglected the energy required to perform inference in the cloud and accounted only for the energy consumed on the \ac{iot} node.

\subsubsection{Battery Lifetime Estimation}
Beyond the experiments, we estimate the achievable battery lifetime for various scenarios, based on the measured energy. We estimate lifetime in hours ($T_{Bat}$) as:

\begin{equation}
T_{Bat} = \frac{C_{Bat} \cdot (1 - SF_{Bat})}{\overline{P}_{Cycle}}
\end{equation}

where $C_{Bat}$ is the battery capacity, $SF_{Bat}$ is a safety factor accounting for self-discharge and the limited usable capacity, and $\overline{P}_{Cycle}$ is the average power over one duty cycle.
We calculate $\overline{P}_{Cycle} $ as follows:

\begin{equation}
\overline{P}_{Cycle} = \frac{E_{Scenario} + E_{Conn}}{T_{Cycle}} + \overline{P}_{Sleep}
\end{equation}

with $E_{Scenario}$ being the reported energy of the active phase (sensing, optional classification, and transmission), and $T_{Cycle}$ being the cycle length in hours. Since the active phase is usually short compared to the sleep period, we approximate the sleep duration by the full cycle length, so that sleep consumes an average power of $\overline{P}_{Sleep}$ over a cycle.
$E_{Conn}$ captures a potential reconnection overhead specific to MQTT scenarios, since the TCP session drops after longer periods of inactivity. In our case, this occurred after roughly 5~min of inactivity; thus, we add a reconnection cost for MQTT scenarios with sleep periods longer than 5~min.
For CoAP and LoRaWAN, and for MQTT cycles shorter than this threshold, we set $E_{Conn} = 0$.

\section{Results \& Discussion}
\label{sec:results-discussion}
Our results provide insights into the classification accuracy of the deployed \acp{cnn} and, more importantly, assess the impact of each (sub-)task of the system on the sensor node's energy consumption. We first illustrate the differences between the tested models and network configurations, and then we use the findings to determine how these manifest across the proposed example use cases and paradigms.

\subsection{Classification Performance}
We quantify the accuracy achievable with the selected \ac{cnn} architectures and datasets under reasonable training constraints and compare it with that of the quantized versions deployed on the tested \acs{mcu}. The results are listed in Table~\ref{tab:model_metrics}.
Fine-tuning of all \acp{cnn} on our domain-specific \acs{plantvillage} dataset led to Top-1 accuracies of $> 99\%$ in all cases. The loss in accuracy due to quantization is small, with less than 0.3\% for \acs{mobilenet} and \acs{squeezenet}  as well as 2.2\% for \acs{mcunet}, which is considered a reasonable trade-off for enabling deployment on resource-constrained \acp{mcu}. In this case, \acs{squeezenet} provides the best trade-off with only 0.79~MB model size, the smallest accuracy loss, and the best performance after quantization.
This is comparable to the accuracy reported in other studies \cite{pandey_survey_2024}.

\begin{table}[h]
\centering
\caption{Model sizes and accuracy metrics. The options with the preferable trade-off between accuracy and model size are emphasized.}
\label{tab:model_metrics}
\resizebox{\columnwidth}{!}{
\begin{tabular}{lllccc}
\cmidrule(lr){1-6}
Data & Model & Type & Size (MB) & Top-1 (\%) & Top-5 (\%) \\
\cmidrule(lr){1-6}
\multirow{6}{*}{PV} & \multirow{2}{*}{MobileNetV2} & Float & 9.38  & {99.55} & {100.00} \\
& & Int & 2.38  & {99.25}& {99.98} \\ \cmidrule(lr){2-6}
& \multirow{2}{*}{SqueezeNet} & Float & 3.00  & {99.51} & {100.00} \\
& & Int & \textbf{0.79}  & \textbf{99.35} & 100.00 \\ \cmidrule(lr){2-6}
& \multirow{2}{*}{MCUNet (in4)} & Float & 6.00  & {99.72} & {100.00} \\
& & Int & 1.6  & {97.56} & {99.95} \\
\cmidrule(lr){1-6}
\multirow{6}{*}{CUB} & \multirow{2}{*}{MobileNetV2} & Float & 10.21  & {74.36} & {91.00}\\
& & Int & \textbf{2.59}  & \textbf{{71.65}} & {90.07} \\ \cmidrule(lr){2-6}
 & \multirow{2}{*}{SqueezeNet} & Float & 3.34  & {66.55} & {88.03}\\
& & Int & 0.87  & {65.53}& {87.44}\\ \cmidrule(lr){2-6}
& \multirow{2}{*}{MCUNet (in4)} & Float & 6.2  & {75.51} & {93.04} \\
& & Int & 1.7  & {70.29} & {90.83} \\
\cmidrule(lr){1-6}
\multirow{2}{*}{YH} & \multirow{2}{*}{TCResNet} & Float & 0.252 & {97.10} & \\
& & Int & \textbf{0.063} & \textbf{{95.87}} &  \\
\cmidrule(lr){1-6}
\end{tabular}}
\end{table}

When trained on the \acs{cub} dataset, the 32-bit models achieved accuracies of 74.36\%~(\acs{mobilenet}), 66.55\%~(\acs{squeezenet}) and 75.51\%~(\acs{mcunet}). Notably, excessive training and larger model architectures can produce models with about 85.6\% accuracy \cite{QSENNCUB}.
Although these values are significantly lower than for the plant disease classification models, the performance is still powerful in light of the challenging dataset covering 200 different classes of bird species, while also considering the small model sizes and limited hyperparameter tuning.
With 71.65\%, 65.53\% and 70.29\% Top-1 accuracy of the quantized \acs{mobilenet}, \acs{squeezenet}, and \acs{mcunet} models, we consider \acs{mobilenet} to be the best option for the \ac{cub} dataset as long as the available hardware is capable of fitting the 2.59~MB quantized model.

\begin{figure*}[t]
    \begin{subfigure}[b]{0.78\textwidth}
        \includegraphics[width=1\linewidth]{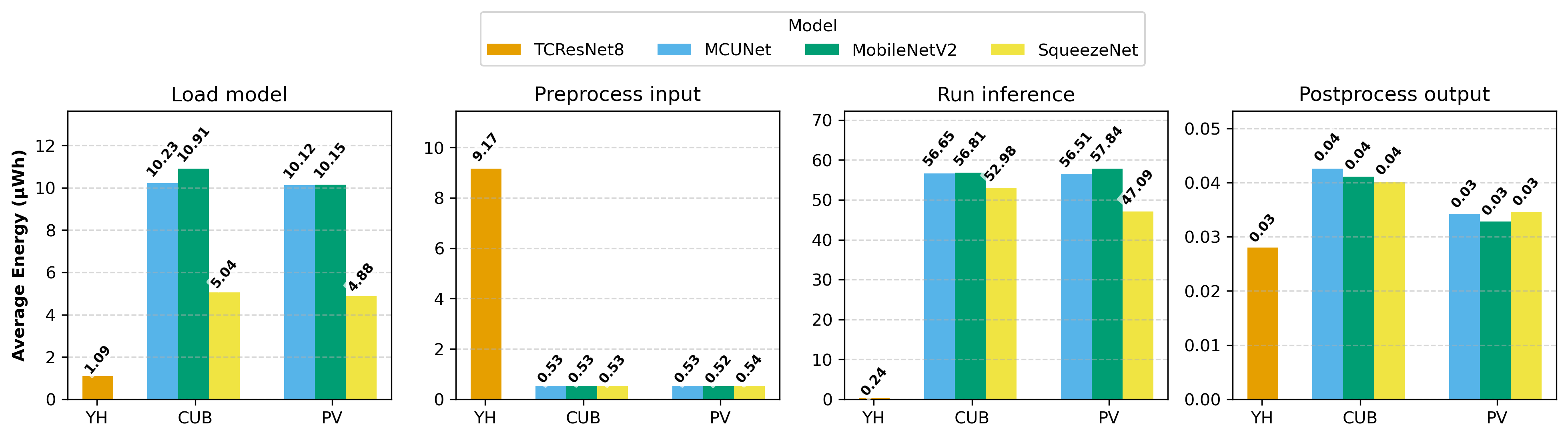}
    \caption{Energy by subtask.}
        \label{fig:amphour-per-inf-a}
        \end{subfigure}%
        \hfill
\begin{subfigure}[b]{0.20\textwidth}
        \includegraphics[width=1\linewidth]{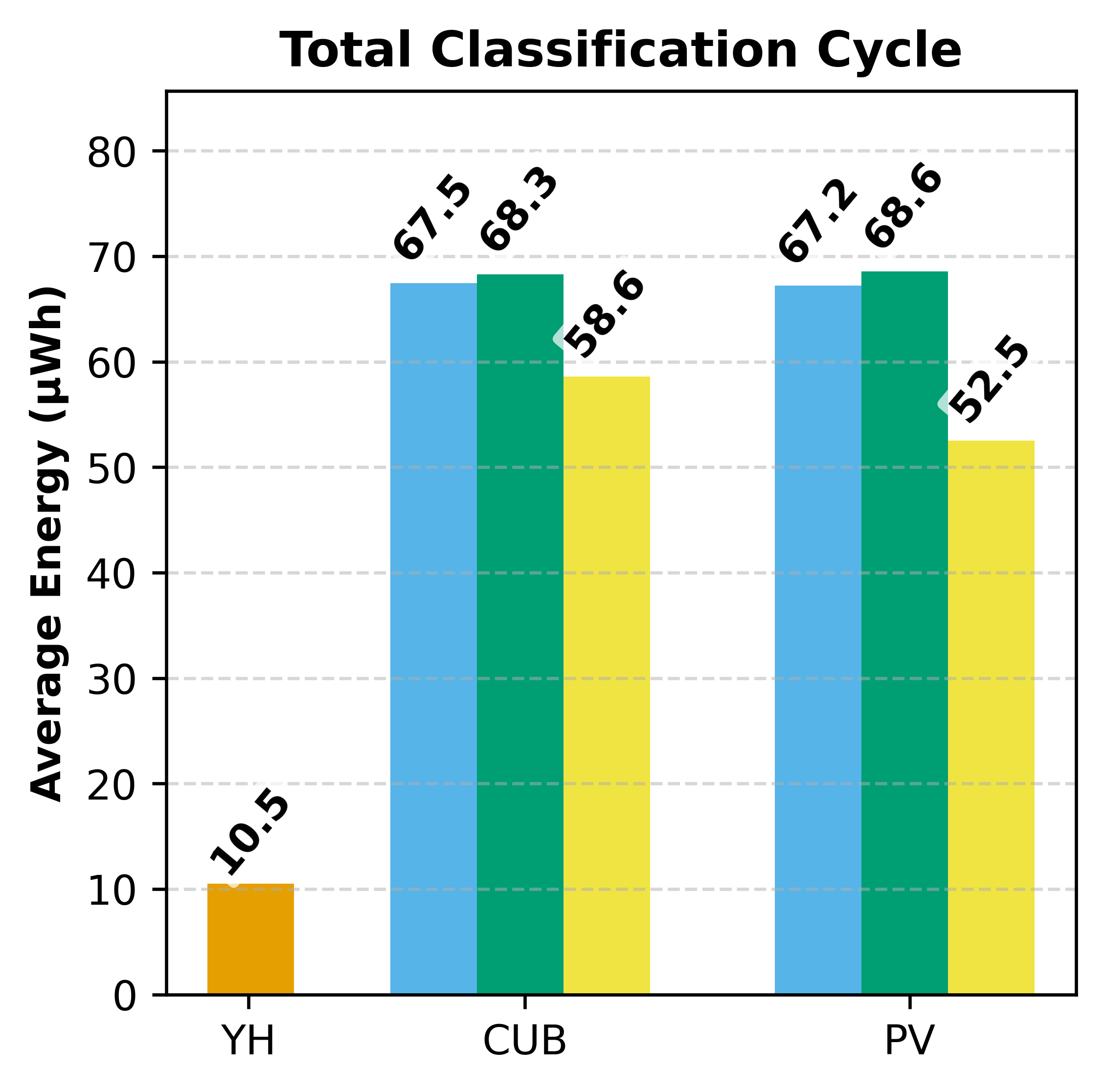}
    \caption{Overall \ac{ml} energy.}
        \label{fig:amphour-per-inf-b}
    \end{subfigure}
    \caption{Average energy footprint of running a single CNN classification cycle across different model architectures and datasets on the test devices. A cycle includes model loading, data preprocessing, inference, and postprocessing.}
    \label{fig:amphour-per-inf}
\end{figure*}

\begin{figure*}[h]
\centering
    \begin{subfigure}[b]{0.7\textwidth}
        \centering
    \includegraphics[width=1\linewidth]{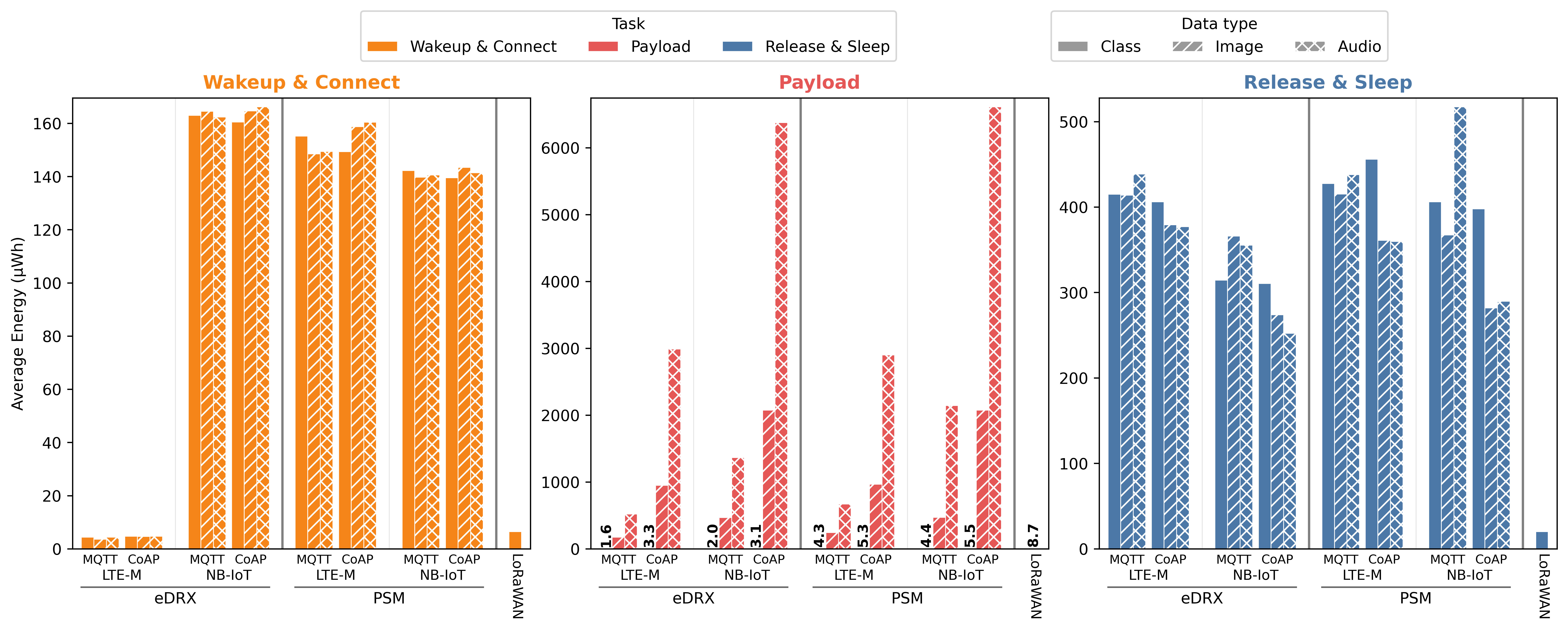}
        \caption{Required energy per sub-task.}

    \end{subfigure}%
    ~ 
    \begin{subfigure}[b]{0.3\textwidth}
        \centering
    \includegraphics[width=1\linewidth]{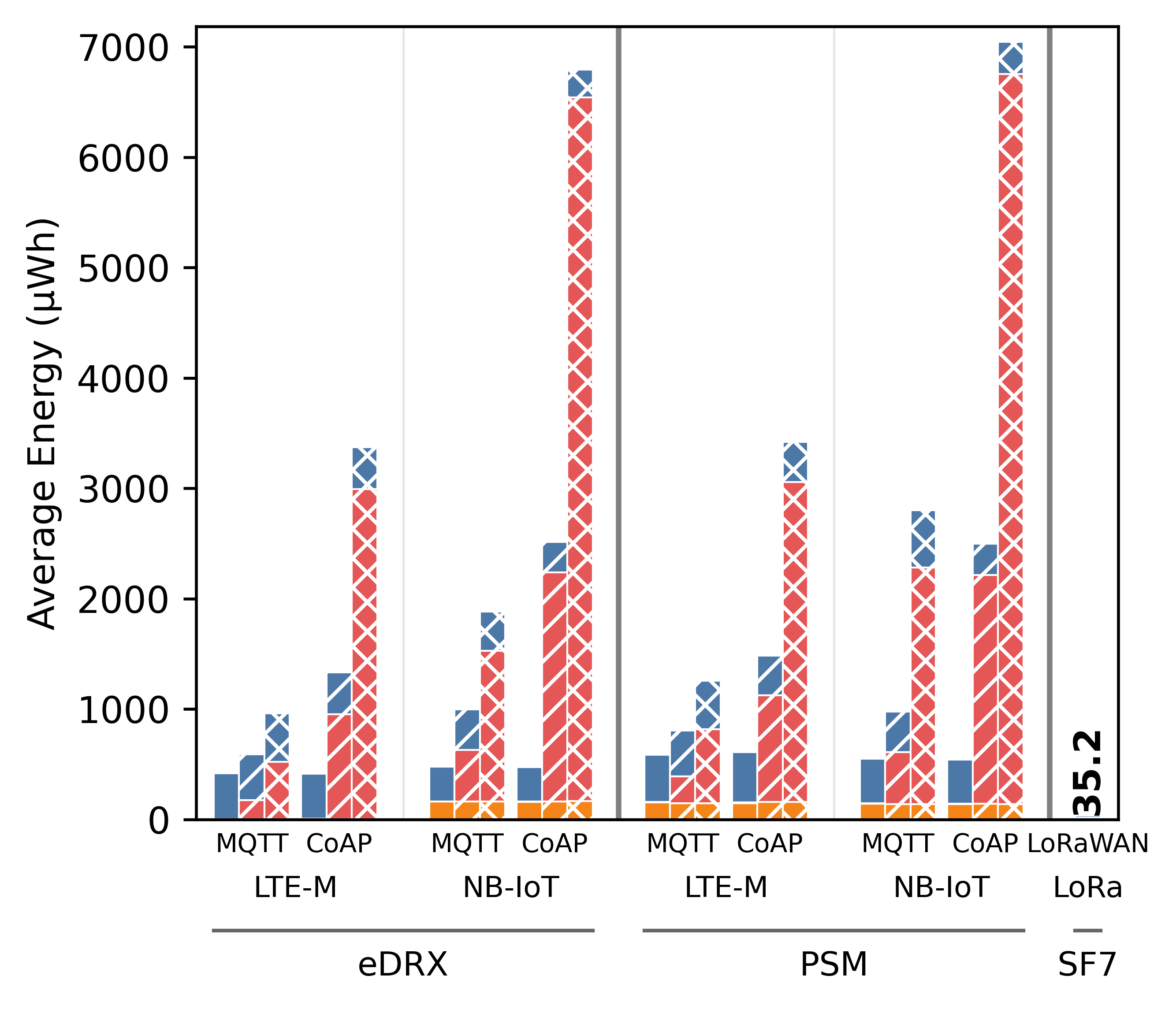}
    \caption{Overall energy per configuration.}
    \end{subfigure}
     \caption{Energy footprint of transmitting class labels, images, and audio using different configurations of networks, protocols, and power-saving modes. Bold text labels were added for small bars that were not visible.} 
    \label{fig:amphour-per-transm}
\end{figure*}

Training the TCResNet model on the \ac{yellowhammer} dataset to detect audio recordings led to an accuracy of 97.10\%. With a small quantization error, the compressed model still detects 95.86\% of the recordings correctly with a model size of only 63~KB.
The memory requirements correspond to about 25\% of the full-precision model size. The compression ratio roughly applies to all models, as the quantization was performed from 32-bit to 8-bit. 
A single epoch of \ac{qat} was sufficient to close the \ac{ptq} gap, delivering final top‑1 accuracies within 0 - 5\% of all 32-bit models.

\subsection{Inference Energy}
To evaluate the energy consumption of the \acp{cnn} inference task, we split a classification cycle into four subtasks (model loading, preprocessing, inference, and postprocessing) as depicted in Figure~\ref{fig:amphour-per-inf-a} and show the overall energy consumption in Figure~\ref{fig:amphour-per-inf-b}.
The model with the lowest memory footprint (63~KB) consumes less than {1.1~$\mu$Wh}for loading and initialization. The medium-sized \acl{squeezenet} (0.79--0.87~MB) consumes about {5~$\mu$Wh}, and the largest models, \acl{mobilenet} and \acl{mcunet} (1.6--2.59~MB), require more than 10 times the power of the smallest model and twice the power of the medium-sized model, at {10.12--10.91~$\mu$Wh}. Notably, the model's memory size does not determine the energy required for loading, as seen with the difference in size but equal energy footprints of \acl{mobilenet} and \acl{mcunet}. The output layer size has a minor impact on energy consumption, with 200-class \acs{cub} models using slightly more energy than 39-class \acs{plantvillage} models across all image classification model architectures. Other potential proxies of loading energy include layer count and activation sizes, which require allocation of differently sized memory buffers, but further investigation is needed to confirm this. Many related works on \ac{tinyml} neglect the initialization phase, treating it as a one-time task. However, IoT devices usually use deep sleep modes, so they must re-initialize each time they wake up. In our proposed use cases, it accounts for around 9--16\% of the inference-related energy.

The preprocessing step of the audio model (\ac{mfcc} creation) accounts for approximately 88\% of its total energy footprint, while inference consumes less than 2.5\% ({0.24~$\mu$Wh}). In contrast, across all image classification models and datasets, inference is the most energy-intensive task, accounting for 83\%--91\% of the total \ac{ml}-related energy.
Notably, inference energy remains stable at {56.8--57.8~$\mu$Wh} for \acs{mobilenet} models and 56.5--56.7~$\mu$Wh for \acs{mcunet} across both datasets. However, SqueezeNet inference varies by dataset, with averages of {47.1~$\mu$Wh} (\acs{plantvillage}) and {53.0~$\mu$Wh} (\acs{cub}).
Postprocessing energy remains marginal across all configurations, ranging from 0.03 to 0.04 $\mu$Wh depending on the output class size (Fig.~\ref{fig:amphour-per-inf-a}). In terms of total energy, \acs{tcresnet} demonstrates the lowest footprint at 10.5 $\mu$Wh. Among the image models, SqueezeNet is the most efficient (52.5--58.6~$\mu$Wh), while MCUNet and MobileNet show comparable performance, consuming 67.2--68.6~$\mu$Wh.

\subsection{Data Transmission Energy}
We tested various configurations of radio technologies (LTE-M, NB-IoT, or LoRaWAN), application protocols (MQTT or CoAP), and power-saving modes (PSM or eDRX) to assess their impact on the energy required to transmit a 1-byte CNN output, a 20 KB image, and a 64 KB audio file. The transmission task comprises sub-tasks of (1) waking up and establishing a network connection, (2) sending the payload, and (3) waiting for the network to release our client and going to sleep until the next cycle starts. Figure~\ref{fig:amphour-per-transm} displays the average energy per sub-task measured across all configurations.
Notably, the eDRX offered an advantage over PSM in the most energy-efficient scenario per message type when using cellular technologies. We therefore report the average energy measured for the eDRX configurations in Table~\ref{tab:amphour-per-transm}, while the PSM configurations are provided in the Supplementary Materials.

\begin{table}[h]
\caption{Energy in $\mu$Wh for the transmission task by different network settings in combination with eDRX. The options with the least energy consumption per data type are emphasized.}
\label{tab:amphour-per-transm}
\resizebox{\columnwidth}{!}{
{\scriptsize
\begin{tabular}{llp{35pt} | p{34pt} p{30pt} p{30pt} | p{20pt}}
\toprule
Data & LPWAN & Protocol / Spreading Factor & Wakeup \& Connect ($\mu$Wh) & Payload ($\mu$Wh) & Release~\& Sleep ($\mu$Wh) & Total ($\mu$Wh) \\
\midrule
\multirow{3}{*}{Class}  & \multirow{2}{*}{LTE-M} & MQTT & \textbf{4.34} & \textbf{1.58} & 414.90 & 420.83 \\
 & & CoAP & 4.73 & 3.33 & 406.04 & 414.10 \\
 & \multirow{2}{*}{NB-IoT} & MQTT & 162.96 & 2.01 & 314.40 & 479.37 \\
\multirow{2}{*}{1 byte} &  & CoAP & 160.51 & 3.13 & 310.45 & 474.10 \\
& LoRaWAN & \ac{sf}7 & 6.45 & 8.65 & \textbf{20.12} & \textbf{35.23}  \\ 
& LoRaWAN & \ac{sf}10 & 5.96 & 35.70 & 24.30 & 65.95  \\ 
& LoRaWAN & \ac{sf}12 & 6.45 & 134.61 & 35.71 & 176.77  \\ \midrule
\multirow{2}{*}{Image}   & \multirow{2}{*}{LTE-M} & MQTT & \textbf{3.60} & \textbf{173.53} & 413.85 & \textbf{590.98} \\
 &  & CoAP & 4.65 & 949.68 & 379.14 & 1333.48 \\
\multirow{2}{*}{20KB} & \multirow{2}{*}{NB-IoT} & MQTT & 164.54 & 466.49 & 366.08 & 997.10 \\
 &  & CoAP & 164.68 & 2074.98 & \textbf{273.85} & 2513.50 \\ \midrule
\multirow{2}{*}{Audio}  & \multirow{2}{*}{LTE-M} & MQTT & \textbf{4.41} & \textbf{519.50} & 438.68 & \textbf{962.59} \\
 &  & CoAP & 4.74 & 2989.28 & 377.03 & 3371.04 \\
\multirow{2}{*}{64KB} & \multirow{2}{*}{NB-IoT}  & MQTT & 162.40 & 1365.90 & 355.35 & 1883.64 \\
 &  & CoAP & 166.21 & 6375.02 & \textbf{252.30} & 6793.53 \\
\bottomrule
\end{tabular}}}
\end{table}

\subsubsection{Class Indices}
Transmitting the class index via LoRaWAN consumes by far the least total amount of energy, with a total of {35.23 to 176.77~$\mu$Wh}. The variation depends on the \ac{sf} that is used for transmission. 
The transmission overhead in LoRaWAN is dominated by the energy required to listen to incoming data from the network using two delayed Rx windows after transmission. 
When considering only the sending cost, without communication overhead, all other protocols are more efficient. The differences between them are almost negligible, with MQTT consuming slightly less energy for the payload and CoAP configurations consuming less energy during release and sleep.
Due to the short sleep interval used in this configuration, we were able to keep the MQTT connection alive during the device sleep phase. Therefore, we avoided the need to reconnect after wakeup, which would add substantial energy overhead in MQTT configurations. Therefore, when MQTT or TCP connections cannot be maintained during device sleep, CoAP is preferable because it runs over connectionless UDP, avoiding the additional power consumption of the TCP handshake. 

For eDRX configurations, we can observe a clear advantage in \textit{Wakeup \& Connect} energy for \mbox{LTE-M} configurations. This difference originates from differences in the mandatory RRC inactivity timer in the chosen networks during our tests and from modem behavior rather than from the power-saving mode itself. In eDRX, the modem chooses a light sleep mode by default, avoiding the need to wait for modem wakeup in the next cycle. This results in roughly {145~$\mu$Wh} less energy consumed per cycle, as seen in the LTE-M configurations. However, when eDRX cycles are longer than 20.48 s, the modem automatically enters deep sleep.
In our tests on Vodafone’s \acs{nbiot} network, the modem’s RRC connection was released after the two-second RRC inactivity timer expired, allowing the modem to spend the remaining $\sim$~28 seconds in eDRX.
Consequently, it switched to deep sleep, requiring more energy to wake up on the next cycle. On Telekom's \acs{ltem} network, the modem needed to wait 10 seconds to release the RRC connection. Therefore, it was not put to deep sleep and woke up faster. As a result, the device consumed more energy during the \textit{Release \& Sleep} phase, but far less energy during the \textit{Wakeup \& Connect} phase.
Thus, eDRX is particularly beneficial for short sleep cycles, as it avoids the need for a complete modem wakeup. For longer sleep cycles, PSM can outperform eDRX, as it consumes slightly less sleep current and allows the device to remain silent for longer periods. We also experienced some network providers restricting the available configurations for both power-saving modes. 

For small payloads, such as the output of a CNN performing a classification task, LoRaWAN is clearly the preferred option. However, when cellular connectivity is required, LTE-M and NB-IoT demonstrated comparable energy efficiency on our device. The observed differences can be attributed primarily to network operator configuration rather than the radio technologies themselves. Therefore, the choice should be guided by the configurations supported by the available operators.

\subsubsection{Sensor Data}
The technical specifications of LoRaWAN limit its suitability to sending only the \ac{cnn} output (applicable only to \texttt{Embedded-ML}). Both cellular radio technologies also allow for sending larger messages. In CoAP, bigger data is transmitted in a block-wise manner, while MQTT may handle larger data packages as a single message at the protocol level.
In contrast to the small energy difference observed with class indices, when larger sensor data (e.g., images or audio files) is transmitted, \ac{nbiot} consumes significantly more energy than \mbox{LTE-M}. This is likely due to higher latencies in \ac{nbiot} and the resulting need for the sensor node to stay online and awake longer while also powering its host \ac{mcu}.
MQTT, combined with LTE-M and eDRX, is the preferred configuration, with an average consumption of {590.98~$\mu$Wh} for a 20~KB image and 962.59~$\mu$Wh for a 64 KB audio file. Sending multiple blocks with CoAP clearly increases the required energy ({1333 $\mu$Wh for images; 3371 $\mu$Wh for audio}). The high energy consumption in CoAP originates from the block-wise transfer mechanism, which splits large messages into 1~KB-sized blocks and waits for confirmation between individual messages. The device's energy efficiency decreases, independent of the application protocol, when switching to \ac{nbiot}. The \acs{mcu} contributes mainly to this.
The \textit{Payload}-related energy scales linearly with more data, while \textit{Wakeup \& Connect} and \textit{Release \& Sleep} energy remain roughly constant. For the exemplary audio file, three times larger than the image, the energy to send the payload approximately triples across all network and application protocol configurations (see Table \ref{tab:amphour-per-transm}).

Overall, MQTT in combination with LTE-M and eDRX is the best choice for sending large messages. However, we also observed a strong dependence on operator-specific network configurations, which will affect the findings, particularly for smaller messages.

\subsection{Sensitivity Analysis}
The sensitivity analysis confirmed the expected negative impact of poor signal conditions on energy efficiency.
In Table~\ref{tab:energy_sa_combined}, we report the mean energy consumption and its standard deviation across varying coverage conditions for the \textit{Payload} transmission sub-task and for the \textit{Total} transmission cycle.
In LoRaWAN, payload energy increases proportionally with a higher \ac{sf} because it directly increases the time-on-air needed to transmit a message. Payload energy increases by a factor of 4.1$\times$ from SF7 to SF10 and by 3.8$\times$ from SF10 to SF12. The total energy scales more moderately (1.9$\times$ and 2.7$\times$, respectively), since the overhead energy is dominated by operations that are independent of the spreading factor, e.g., waiting for RX windows.

\begin{figure}[h]
    \centering
    \begin{subfigure}[t]{0.48\columnwidth}
        \centering
        \includegraphics[width=\linewidth]{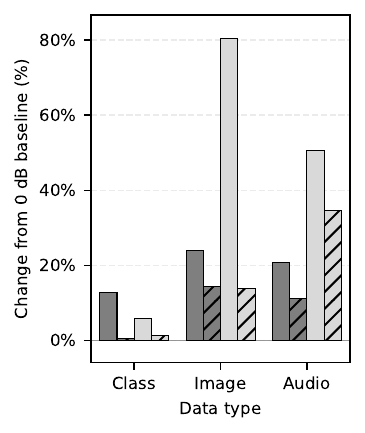}
        \caption{Payload energy}
        \label{fig:energy-payload}
    \end{subfigure}
    \hfill
    \begin{subfigure}[t]{0.48\columnwidth}
        \centering
        \includegraphics[width=\linewidth]{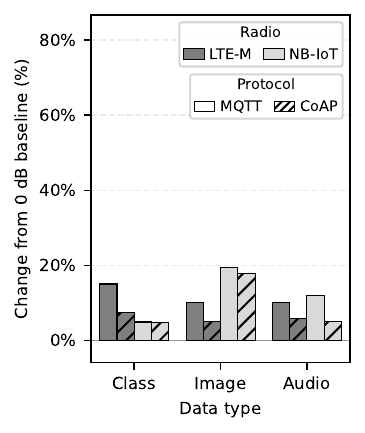}
        \caption{Overhead energy}
        \label{fig:energy-overhead}
    \end{subfigure}
    \caption{
        Relative energy change at 30 dB attenuation compared with the 0 dB baseline.
        Bar fill encodes the radio technology, while hatching encodes the application protocol.
    }
    \label{fig:energy-relative-30db}
\end{figure}

For cellular networks, we observe a significant increase in energy consumption only under severe attenuation of 30 dB. Notably, the \ac{nbiot} results have a higher standard deviation, which can produce misleading results. For instance, the energy required to send an image via CoAP decreases from 0 to 15 dB. However, this is an exception, and the energy required generally increases or remains about constant for the smaller attenuation level of 15 dB. LTE-M was the more stable and preferable option overall. The highlighted numbers show the preferred application protocol per attenuation level and network technology.
Figure~\ref{fig:energy-relative-30db} demonstrates the relative increase in energy consumption across cellular network configurations under 30 dB link attenuation, compared to an unattenuated (0 dB) link.
Hereby, we distinguish between the energy of the payload sub-task (Figure \ref{fig:energy-payload}) and the overhead energy (Figure \ref{fig:energy-overhead}), including both the \textit{Wakeup \& Connect} and \textit{Release \& Sleep} sub-tasks. 
The impact of poor signal conditions on overhead energy ranges from 7\% to 20\% across all data types (i.e., data sizes). Meanwhile, it can be seen that the impact on payload energy scales higher for large-sized payloads: 2\%--16\% for a one-byte-sized payload (class) compared to an increase of 13\%--80\% for a 20 KB payload (image), and 11\%--50\% for a 64 KB payload (audio).

Overall, LoRaWAN remains the more energy-efficient option compared to cellular, even under unfavorable conditions. At SF12, its most energy-intensive setting, LoRa still consumes less than half the energy of the most favorable cellular network under strong signal conditions without attenuation.
We refrain from further direct comparisons between \ac{sf} values and attenuation levels, as they represent fundamentally different physical parameters and thus are not directly comparable.

\begin{table}[t]
\centering
\caption{
Sensitivity of transmission-related energy to varying coverage conditions. For cellular networks, the most efficient option per condition is highlighted in bold.}
\label{tab:energy_sa_combined}
\resizebox{\columnwidth}{!}{
\begin{tabular}{l l c c c c | c c c c}
\toprule
\textbf{Data} & 
\multicolumn{2}{l}{\textbf{Condition}} & 
\multicolumn{3}{l|}{\textbf{Payload} ($\mu$Wh)} & 
\multicolumn{4}{c}{\textbf{Total} ($\mu$Wh)} \\
\\[-1.8ex]
\midrule
\\

\multicolumn{6}{l|}{\textit{\textbf{LoRaWAN}} (spreading factor)} & \multicolumn{4}{l}{} \\
\cmidrule(lr){1-10}
& \shortstack[l]{\textbf{SF}} & \multicolumn{2}{c}{Mean} & \multicolumn{2}{c|}{SD} 
  & \multicolumn{2}{c}{Mean} & \multicolumn{2}{c}{SD} \\
\cmidrule(lr){1-10}
\multirow{3}{*}{\textbf{Class}} & SF7  & \multicolumn{2}{c}{8.65}   & \multicolumn{2}{c|}{0.03} & \multicolumn{2}{c}{35.23} & \multicolumn{2}{c}{0.36} \\
                                & SF10 & \multicolumn{2}{c}{35.70}  & \multicolumn{2}{c|}{0.08} & \multicolumn{2}{c}{65.95} & \multicolumn{2}{c}{0.26} \\
                                & SF12 & \multicolumn{2}{c}{134.61} & \multicolumn{2}{c|}{6.01} & \multicolumn{2}{c}{176.77} & \multicolumn{2}{c}{6.33} \\
                                
\midrule
\\
\multicolumn{6}{l|}{\textit{\textbf{LTE-M}} (attenuation level)} & \multicolumn{4}{l}{} \\
\cmidrule(lr){1-10}
& \multirow{2}{*}{\shortstack[l]{\textbf{Atten.}}} & \multicolumn{2}{c}{\textbf{MQTT}} & \multicolumn{2}{c|}{\textbf{CoAP}} 
  & \multicolumn{2}{c}{\textbf{MQTT}} & \multicolumn{2}{c}{\textbf{CoAP}} \\
\cmidrule(lr){3-4} \cmidrule(lr){5-6} \cmidrule(lr){7-8} \cmidrule(lr){9-10}
& & Mean & SD & Mean & SD & Mean & SD & Mean & SD \\
\cmidrule(lr){1-10}
\multirow{3}{*}{\textbf{Class}} & 0 dB & 2.16 & 0.15 & 3.16 & 0.28 & \textbf{391.77} & 58.06 & 395.76 & 21.28 \\
& 15 dB & 2.21 & 0.05 & 3.16 & 0.18 & 413.26 & 44.69 & \textbf{396.97} & 13.88 \\
& 30 dB & 2.44 & 1.06 & 3.18 & 0.08 & 450.82 & 43.23 & \textbf{425.34} & 27.43 \\
\cmidrule(lr){1-10}
\multirow{3}{*}{\textbf{Image}} & 0 dB & 166.53 & 6.80 & 781.99 & 68.05 & \textbf{578.74} & 17.93 & 1172.77 & 72.60 \\
& 15 dB & 177.00 & 14.02 & 804.11 & 65.28 & \textbf{601.60} & 21.86 & 1192.28 & 66.29 \\
& 30 dB & 206.51 & 5.71 & 895.70 & 54.68 & \textbf{661.03} & 24.20 & 1306.04 & 65.87 \\
\cmidrule(lr){1-10}
\multirow{3}{*}{\textbf{Audio}} & 0 dB & 508.63 & 22.67 & 2440.52 & 222.92 & \textbf{934.55} & 34.54 & 2818.08 & 220.87 \\
& 15 dB & 522.39 & 48.29 & 2490.28 & 159.36 & \textbf{955.30} & 49.81 & 2873.32 & 160.26 \\
& 30 dB & 614.26 & 26.41 & 2715.30 & 168.70 & \textbf{1083.28} & 36.88 & 3115.58 & 176.50 \\
\midrule
\\

\multicolumn{6}{l|}{\textit{\textbf{NB-IoT}} (attenuation level)} & \multicolumn{4}{l}{} \\
\cmidrule(lr){1-10}
& \multirow{2}{*}{\shortstack[l]{\textbf{Atten.}}} & \multicolumn{2}{c}{\textbf{MQTT}} & \multicolumn{2}{c|}{\textbf{CoAP}} 
  & \multicolumn{2}{c}{\textbf{MQTT}} & \multicolumn{2}{c}{\textbf{CoAP}} \\
\cmidrule(lr){3-4} \cmidrule(lr){5-6} \cmidrule(lr){7-8} \cmidrule(lr){9-10}
& & Mean & SD & Mean & SD & Mean & SD & Mean & SD \\
\cmidrule(lr){1-10}
\multirow{3}{*}{\textbf{Class}} & 0 dB & 1.95 & 0.56 & 2.79 & 0.16 & 537.35 & 97.24 & \textbf{499.17} & 88.14 \\
& 15 dB & 1.83 & 0.06 & 2.86 & 0.16 & 494.96 & 52.00 & \textbf{486.57} & 73.50 \\
& 30 dB & 2.06 & 0.73 & 2.82 & 0.17 & 564.08 & 205.11 & \textbf{523.30} & 60.99 \\
\cmidrule(lr){1-10}
\multirow{3}{*}{\textbf{Image}} & 0 dB & 416.74 & 47.26 & 1899.68 & 165.67 & \textbf{910.53} & 59.01 & 2312.20 & 176.13 \\
& 15 dB & 481.24 & 38.28 & 1798.93 & 309.66 & \textbf{1009.71} & 57.01 & 2220.78 & 316.89 \\
& 30 dB & 751.43 & 247.49 & 2162.89 & 291.55 & \textbf{1341.10} & 325.38 & 2649.10 & 432.37 \\
\cmidrule(lr){1-10}
\multirow{3}{*}{\textbf{Audio}} & 0 dB & 1413.27 & 168.86 & 5668.85 & 802.32 & \textbf{1939.96} & 197.86 & 6089.86 & 819.81 \\
& 15 dB & 1626.03 & 54.20 & 6320.56 & 1001.90 & \textbf{2164.94} & 65.67 & 6754.95 & 996.91 \\
& 30 dB & 2129.72 & 118.04 & 7637.87 & 644.02 & \textbf{2720.06} & 113.12 & 8080.14 & 670.34 \\

\bottomrule
\end{tabular}
}
\end{table}
 
\subsection{Application Energy}
In addition to the energy measurements from the previous experiments, we directly measured the sensing and sleep tasks. For the sensing task, we measured an average of 31.23~$\mu$Wh to take a picture with the OV2640 camera and 76.97~$\mu$Wh to record a two-second audio snippet with the MEMS microphone. For the sleep task, we report the average power of 37.52~$\mu$W rather than energy, since sleep duration varies across use cases. We combined the measurements with the reported values from the experiments to evaluate energy consumption across the paradigms and modes illustrated in Figure~\ref{fig:paradigms} and use cases described earlier (see Section \ref{subsec:datasets}). In this final analysis, we address two aspects: 
(1) We compare the impact of the different paradigms and modes on the energy consumption of a single application cycle (with a 30 s sleep interval) across each use case, and (2) we estimate the battery lifetime of each application for varying duty cycles (i.e., sleep interval durations).

\subsubsection{Per-Cycle Energy}
\begin{figure}[H]
    \centering
    \includegraphics[width=1\linewidth]{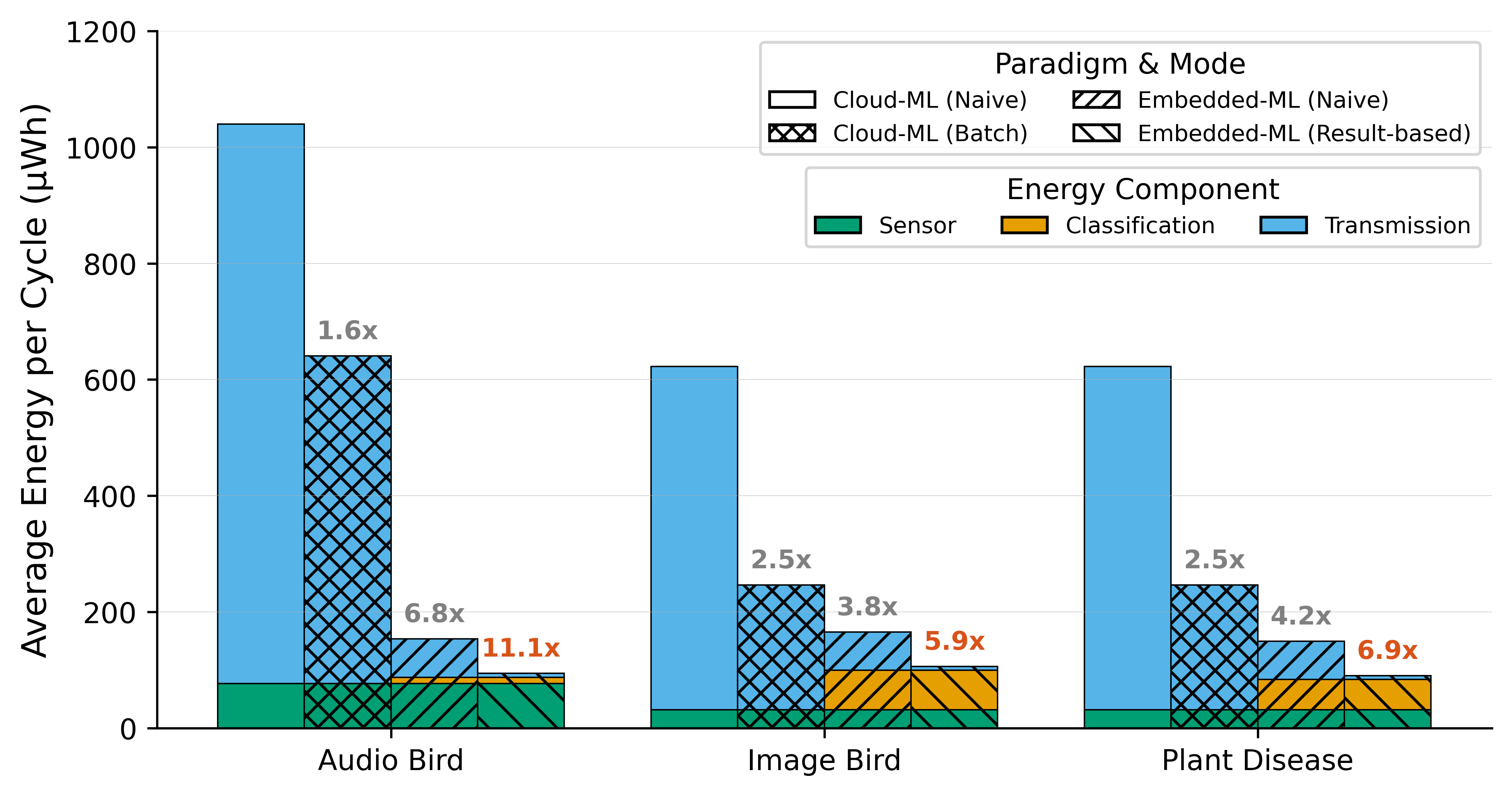}
    \caption{Average energy per cycle (application + 30\,s sleep) for different paradigms and modes across three use cases, decomposed by sensor, classification, and transmission costs. \texttt{Cloud-ML} and \texttt{Embedded-ML} scenarios transmit via LTE-M/MQTT and LoRaWAN \ac{sf}10, respectively. Local inference for \texttt{Embedded-ML} is performed using the best energy-accuracy tradeoff models: TC-ResNet8 (\texttt{Audio Bird}), MobileNetV2 (\texttt{Image Bird}), and SqueezeNet (\texttt{Plant Disease}). Values above the bars represent the energy decrease factor compared to the \texttt{Cloud-ML} naive baseline.}
    \label{fig:full-cycle-energy}
\end{figure}

For all scenarios, we considered sensor energy (camera or microphone) and a full transmission cycle (wake up \& connect, sending, and overhead). In addition, for \texttt{Embedded-ML} scenarios, we added the energy of a full classification cycle (model loading, preprocessing, inference, and postprocessing) from the corresponding \ac{cnn}. The results are depicted in Figure \ref{fig:full-cycle-energy}. 
From Table \ref{tab:amphour-per-transm}, the most energy-efficient network option for transmitting a 20~KB image and a 64~KB audio file was derived with LTE-M and MQTT in both cases. For transmission of the \ac{cnn} output, LoRaWAN proved to be the best option. In LoRaWAN, the spreading factor is a design parameter, since we can deploy our own gateways. However, achieving low spreading factors like SF7 requires close-range gateways and thus extra deployment effort. We therefore adopted the longer-ranging SF10 for the per-cycle energy comparison, reducing the deployment requirements and avoiding overly favoring LoRaWAN.
The required transmission frequency for the result-based mode of the Embedded-ML paradigm is highly use-case specific and cannot be generalized. To demonstrate the effect of this mode, we consider an exemplary scenario in which data is transmitted after every tenth cycle, resulting in an average transmission energy that is one-tenth of the naive mode's transmission energy.
To further strengthen the comparison, we also demonstrate an alternative mode of \texttt{Cloud-ML}, in which a batch of ten sensor measurements is sent every tenth cycle. In this mode, we consider the average transmission energy overhead to be one-tenth of the \texttt{Cloud-ML} scenario, while average payload transmission energy remains constant. 

For each use case, the \texttt{Cloud-ML} naive scenario consumes the most energy, at 1039.56~$\mu$Wh for \textit{Audio Bird} and 622.21~$\mu$Wh for the image use cases. To evaluate the potential energy gains of simple data batching, we observed that reducing transmission frequency reduces energy consumption by a factor of 1.6$\times$ (640.78~$\mu$Wh for \textit{Audio Bird}) to 2.5$\times$ (246.50~$\mu$Wh for image use cases). However, this approach introduces significant system response latency, and even for the individually short 30~s sleep intervals, accumulating radio silence risks triggering TCP timeouts when using the MQTT protocol and requiring additional energy for MQTT reconnection. Consequently, the reported batching gains represent a simplified best-case scenario and may be less achievable in practice.
In contrast, the \texttt{Embedded-ML} paradigm demonstrates superior energy efficiency without the same latency trade-offs and allows for indefinitely long sessions with LoRaWAN. 
Energy consumption is reduced by $6.8\times$ for \texttt{Audio Bird}, $3.8\times$ for \texttt{Image Bird}, and $4.2\times$ for \texttt{Plant Disease} compared with the respective \texttt{Cloud-ML} naive baseline. At the same time, the quantized models incur only $1.23$, $2.71$, and $0.16$ \% Top-1 accuracy loss, respectively, relative to their full-precision counterparts. This trade-off improves further in the result-based mode of the \texttt{Embedded-ML} paradigm. Assuming an exemplary transmission frequency of 10\% of all measurements, energy consumption decreases to 94.06~$\mu$Wh, 106.12~$\mu$Wh, and 90.33~$\mu$Wh, corresponding to 11.1$\times$ for \texttt{Audio Bird}, 5.9$\times$ for \texttt{Image Bird}, and 6.9$\times$~for~\texttt{Plant~Disease}.

\subsubsection{Battery Lifetime}
\begin{figure}[t]
    \centering
    \includegraphics[width=1\linewidth]{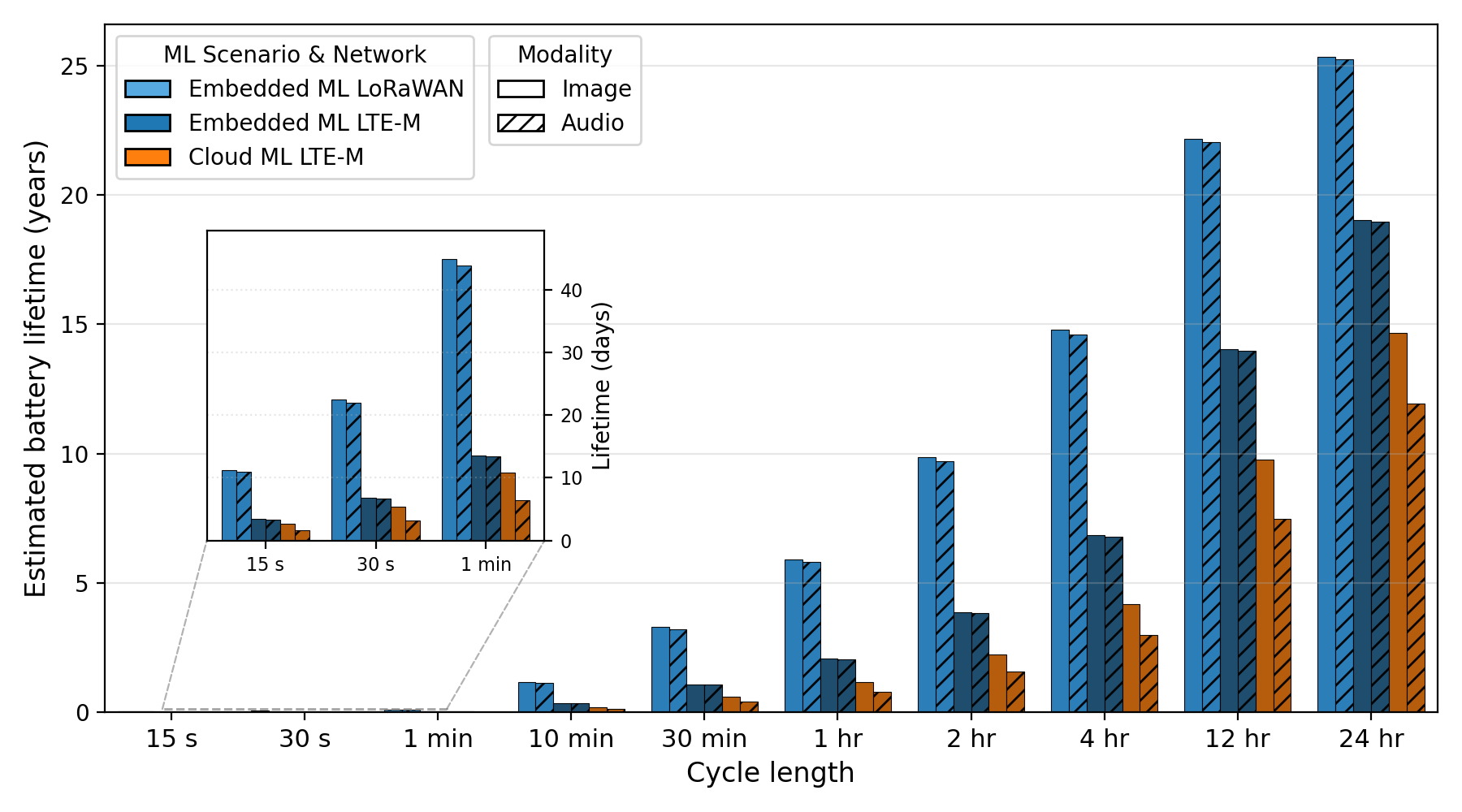}
    \caption{Battery lifetime with different transmission cycles from 10 minutes to once a day. A 12.95~Wh battery with a self-discharge factor of 0.75\% is assumed. In all scenarios, the \texttt{Embedded-ML} is the preferable option.}
    \label{fig:battery-lifetime}
\end{figure}

For our estimation of the battery lifetime, we assumed a sensor node to be powered by a standard Li-Ion 18650 battery with $C_{Bat} =12.95~Wh$ (3500 mAh), and chose a safety factor of $SF_{Bat} = 0.25$ (comparable to \cite{labdaoui2023energy}). We used the Walter board to measure the average deep sleep power $\overline{P}_{Sleep} = 37.52~\mu$W and an average $E_{Conn} = 288.51~\mu$Wh for reconnection in MQTT configurations. We calculated the battery lifetime estimation based on these values for various scenarios and transmission frequencies and report the results in Figure~\ref{fig:battery-lifetime}. 
We can observe that while absolute battery lifetime gain increases with longer sleep intervals, the relative efficiency of the \texttt{Embedded-ML} paradigm is the highest for high-frequency cycles.

For plant disease detection, where images are processed once daily, the system achieves maximum lifetime. In the \texttt{Embedded-ML} scenario, the estimated lifetime reaches up to 25.33 years. Even with less efficient LTE-M connectivity, lifetimes remain high (19.02 years for \texttt{Embedded-ML} vs. 14.68 years for \texttt{Cloud-ML}), suggesting that for low-frequency applications, sleep duration becomes more critical than the choice of radio and paradigm. However, the absolute improvements through the use of \texttt{Embedded-ML} of 10.65 years (LoRaWAN) and 4.34 years (LTE-M) may still justify its use for low-frequency applications. 

The bird feeder scenario is event-driven, as external triggers cause unpredictable wake-up intervals. In such cases, transmission energy cost becomes the dominant factor. For an image-based system with a 1-hour cycle, switching from \texttt{Cloud-ML} to \texttt{Embedded ML} via LoRa increases battery life from 1.17 to 5.93 years (a 5.07$\times$ increase). This demonstrates that embedded processing is essential for maintaining multi-year operation in comparable environments.

The most demanding case is audio-based bird classification. When considering a noise-triggered average interval of 15 seconds, the \texttt{Cloud-ML} baseline lasts only 1.6 days, while the \texttt{Embedded-ML} configuration extends this to 11.0 days (nearly 7$\times$ improvement). When the device operates only 15 minutes per hour, as is often the case for a bio-acoustics data logger, the embedded approach provides 44 days of lifetime compared to only 6.5 days for Cloud-ML. This highlights that Embedded ML may be necessary to make certain high-frequency monitoring applications feasible.

\section{Limitations \& Outlook}\label{sec:limitations_outlook}
A clear limitation of processing sensor data on edge devices, rather than sending it to a server, is the loss of raw data for potential future review or analysis. It limits the applicability to cases where further examination or verification of the raw data is not required. 
However, from a data privacy perspective, not storing or sending any raw data can be beneficial, as it prevents the exposure of sensitive information while still allowing useful analysis based on processed data.

The model architectures deployable on the \ac{mcu} are inherently constrained by memory limits. To isolate the energy trade-off from server-side variables (such as capacity and cost), we restricted our comparison to quantized models that fit on the \ac{mcu}. 
Despite high accuracy in simple tasks (e.g., plant disease detection and bird sound classification at $>$ 99\%), a performance gap likely exists between these compact models and larger server-side alternatives in more complex use cases. Consequently, the applicability of the Embedded-ML approach depends on whether adequately sized models can address the use case with acceptable accuracy. However, as computational efficiency and operator support improve, the range of compatible models is expected to expand, mitigating this limitation.
Beyond compressing existing architectures, Neural Architecture Search (NAS) offers another viable option by enabling the systematic design and optimization of MCU‑specific models for energy efficiency, albeit at the cost of substantially increased training time and reduced compatibility. This is reflected in our MCUNet results, where the gains seen on ARM architectures \cite{linMCUNetTinyDeep2020a} did not translate to a significant improvement in the energy-accuracy trade-off on the Xtensa-based ESP32S3 \ac{mcu}.

Notably, classification is merely one of many applications when processing sensor data. While we expect energy consumption to stay stable for models with similar size and number of operations, applications with larger output tensors will naturally have higher communication costs, reducing the potential for improved energy efficiency. This holds especially true for image segmentation tasks, where one can not estimate the image segment to be sent. Those applications might still benefit from result-based sending. Regression, multi-label classification, and object detection applications are likely to profit similarly to the presented classification applications based on comparably small output sizes. 

Our work is also limited by network operator-controlled configurations. The overhead of cellular networks largely depends on settings that are decided by the network operator. In our tests, we observed crucial differences between available providers, limiting the availability and effectiveness of power-saving options. 
For TCP-based protocols such as MQTT, we maintained active connections with the targeted server over lengthy periods of time to avoid energy-costly reconnection handshakes after deep sleep. However, in real-world deployments, TCP timeouts, unstable network conditions, or server-side session limits can lead to half-open connections and require reconnection tries, degrading the overall energy efficiency of the application.
The present results may not generalize to mobile deployments characterized by fluctuating network conditions and delayed response times.
We limited our selection of \ac{lpwan} to three widely used technologies in LTE-M, NB-IoT, and LoRaWAN. However, emerging technologies like 5G NR RedCap and Wi-Fi HaLow might be beneficial when larger messages need to be sent, given their higher throughput and the resulting need to keep the host \ac{mcu} active for shorter periods.
We expect BLE Long Range to be slightly more energy-efficient than LoRaWAN, but with a shorter range. As such, it is suited where mid-range coverage is sufficient and high efficiency is required.

\section{Conclusion}\label{sec:conclusion}
Energy efficiency is an important metric across most \ac{iot} applications; it becomes crucial for remotely positioned devices that must operate for long periods without maintenance and have limited available energy capacity. Using embedded processing, especially \ac{tinyml}, can heavily reduce the amount of data that needs to be transmitted through costly wireless communication while preserving the relevant bits of information. Processing on the embedded device adds energy overhead while also enabling more lightweight communication technologies.
We provided a systematic evaluation of task-based energy consumption within \ac{tinyml} empowered \ac{iot} deployments. Rather than investigating isolated components, we analyzed the interplay between choice of model architecture, communication protocols, and transmission strategies to identify and quantify the most energy-efficient configurations in operational applications.
Our results show that only certain architectures are suitable for deployment on resource-constrained \acp{mcu} through quantization. However, these architectures experienced minimal accuracy loss.
Furthermore, the evaluation of communication protocols, including different use cases, message sizes, application layer protocols, and power-saving configurations, provides transferable insights that can be applied to a range of similar \ac{iot} applications. 
We demonstrated that performing inference with embedded \acp{cnn} for classification on \acp{mcu} like the ESP32-S3 can significantly reduce the energy consumption of sensor nodes in environmental monitoring applications, especially through the use of more lightweight communication technologies. 
By reducing high-dimensional image or audio data to a single 8-bit classification output through on-board inference, we achieve a significant theoretical battery-lifetime increase of $5\times$ and $6.8\times$ for the proposed visual- and sound-based bird monitoring use cases, respectively. Leveraging inference results to decide whether data transmission is necessary can further improve energy savings. These savings promote deployments in remote locations that lack broadband connectivity by making them compatible with low-bandwidth \ac{lpwan} protocols such as \ac{nbiot}, LTE-M, and LoRaWAN.
By taking advantage of \texttt{\ac{EmML}}, \ac{iot} deployments can achieve greater sustainability and autonomy, ensuring long-term operation in resource-constrained environments while significantly reducing their energy footprint.

\bibliographystyle{IEEEtran}

\end{document}